\documentclass{article}

\usepackage[main, final]{neurips_2026}

\usepackage[utf8]{inputenc} % allow utf-8 input
\usepackage[T1]{fontenc}    % use 8-bit T1 fonts
\usepackage{hyperref}       % hyperlinks
\usepackage{url}            % simple URL typesetting
\usepackage{booktabs}       % professional-quality tables
\usepackage{amsfonts}       % blackboard math symbols
\usepackage{nicefrac}       % compact symbols for 1/2, etc.
\usepackage{microtype}      % microtypography
\usepackage{xcolor}         % colors

\usepackage{wrapfig}
\usepackage{booktabs}
\usepackage{subfigure}
\usepackage{enumitem}
\usepackage{rotating}
\usepackage{multirow}
\usepackage{booktabs,bm}
\usepackage{algorithm}
\usepackage{graphicx}
\usepackage{mathtools}
\usepackage{subcaption}
\usepackage{mdwlist}
\usepackage{amsmath}
\usepackage{amssymb}
\usepackage{amsthm}
\usepackage{hyperref}
\usepackage{colortbl}
\usepackage{pifont}   % \ding{}
\usepackage{xspace}

\usepackage{bm} % 如果使用了 \bm

\usepackage{algorithm}
\usepackage{algpseudocode}
\usepackage{amsmath}

\usepackage{fontawesome}

\definecolor{bestColor}{HTML}{faecf7}   
\definecolor{secondColor}{HTML}{e4f3fa}
\definecolor{worstColor}{HTML}{D9EAD3}  

\newcommand{\best}[1]{\cellcolor{bestColor}\textbf{#1}}
\newcommand{\second}[1]{\cellcolor{secondColor}#1}

\usepackage[dvipsnames]{xcolor}

\hypersetup{
    colorlinks=true,
    linkcolor=teal,  
    citecolor=MidnightBlue,   
    filecolor=magenta,      
    urlcolor=MidnightBlue,
    }

\definecolor{background_gray}{gray}{0.92}

\newcommand{\ourmethod}{{\fontfamily{lmtt}\selectfont \textbf{Tyche}}\xspace}
\title{Tyche: One Step Flow for Efficient Probabilistic Weather Forecasting}

\author{%
  \small Fan Xu$^{1,2*}$, Yuan Gao$^{3*}$, Kun Wang$^{4}$, Rui Su$^{2 \dag}$, Fenghua Ling$^{5}$,  Hao Wu$^{3 \dag}$, Wanli Ouyang$^{2,6 \dag}$ \\
  \small $^{1}$ University of Science and Technology of China  $^{2}$ Shenzhen Loop Area Institute  $^{3}$ Tsinghua University  \\
  \small $^{4}$ Nanyang Technological University  $^{5}$ Shanghai AI Lab  $^{6}$ The Chinese University of Hong Kong \\ 
  \small	 $^*$ Equal Contribution,  $^\dag$ Corresponding author \\
  \small  {\faEnvelope} {Main Contact}: \texttt{markxu@mail.ustc.edu.cn}
}

\begin{document}

\maketitle

\begin{abstract}

Probabilistic weather forecasting requires not only accurate trajectories, but calibrated distributions over plausible atmospheric futures. Recent data-driven systems have achieved remarkable deterministic skill, and diffusion-based ensemble forecasters have substantially improved sample realism and uncertainty quantification. However, their inference cost scales with forecast horizon, ensemble size, and the number of denoising steps required for each transition, making large operational ensembles expensive. To address this, we present \ourmethod, a one-step conditional flow model for efficient probabilistic weather forecasting. \ourmethod models the conditional forecast distribution with a destination-aware average-velocity flow that maps Gaussian noise directly to future weather states in a single function evaluation (1-NFE). To make this one-step transport learnable in high-dimensional geophysical fields, we derive a JVP-regularized rectification objective that enforces temporal self-consistency across source and destination flow timesteps without explicitly forming Jacobians. The transport field is parameterized by an isotropic Swin-style transformer that preserves fine-scale spatial structure while remaining scalable on global grids. To improve ensemble reliability under autoregressive forecasting, we further introduce a rollout-based finetuning stage with curriculum CRPS calibration supervision. Experiments on ERA5 at 1.5$^\circ$ and 6-hour resolution show that our \ourmethod, using merely a single NFE, matches or exceeds the forecast skill and calibration of state-of-the-art multi-step generative baselines and the operational ECMWF IFS ensemble. \href{https://github.com/Markus-xu/Tyche}{Codes are available.}

\end{abstract}

\section{Introduction}
\label{introduction}

Many high-impact decisions in aviation, energy, agriculture, disaster response, and climate-risk management~\citep{mcphillips2018defining,meenal2022weather,raymond2020understanding,wu2025advanced} depend not on a single most likely atmospheric trajectory, but on a calibrated distribution over plausible futures. This is especially important at medium-range lead times, where small perturbations in the initial state can amplify into substantially different outcomes. Recent deterministic machine-learning weather prediction (MLWP) systems~\citep{alet2025skillful,cui2025forecasting,grover2015deep,wang2025gen2,gao2026neuralom,wu2026pnpcorrector}, such as FourCastNet~\citep{pathak2022fourcastnet}, Pangu-Weather~\citep{bi2023accurate}, GraphCast~\citep{lam2023learning}, and NeuralGCM~\citep{kochkov2024neural}, have achieved comparable or better performance compared to traditional operational weather prediction systems~\citep{brotzge2023challenges,nipen2020adopting,ren2021deep}. Although they have much lower computing costs, these deterministic models fail to faithfully represent high-impact extreme events as forecast uncertainty grows over time.

Recent work on global probabilistic weather forecasting has begun to close this gap. One line of work revisits the forecast objective itself, replacing deterministic regression with explicitly probabilistic training, for example through hierarchical latent-variable graph models~\citep{oskarsson2024probabilistic,xu2026advanced} or proper-score optimization in AIFS-CRPS~\citep{lang2026aifs}. Another line embraces explicit generative modeling. Diffusion-based systems~\citep{karras2022elucidating} such as GenCast produce sharp samples with realistic spectra and strong extreme-event skill~\citep{price2025probabilistic}. Follow-up work explores direct lead-time conditioning with temporally correlated noise~\citep{andrae2025continuous}, rolling diffusion with progressive uncertainty schedules~\citep{cachay2025elucidated}, latent-space diffusion~\citep{zhuang2025ladcast}, and one-step consistency-style forecasting~\citep{stock2025swift}. Together, these works establish that ensemble skill, calibration, and realism are all attainable in data-driven forecasting. 

\begin{wrapfigure}[15]{r}{0.5\linewidth}
    \centering
    \vspace{-15pt}
    \includegraphics[width=1.0\linewidth]{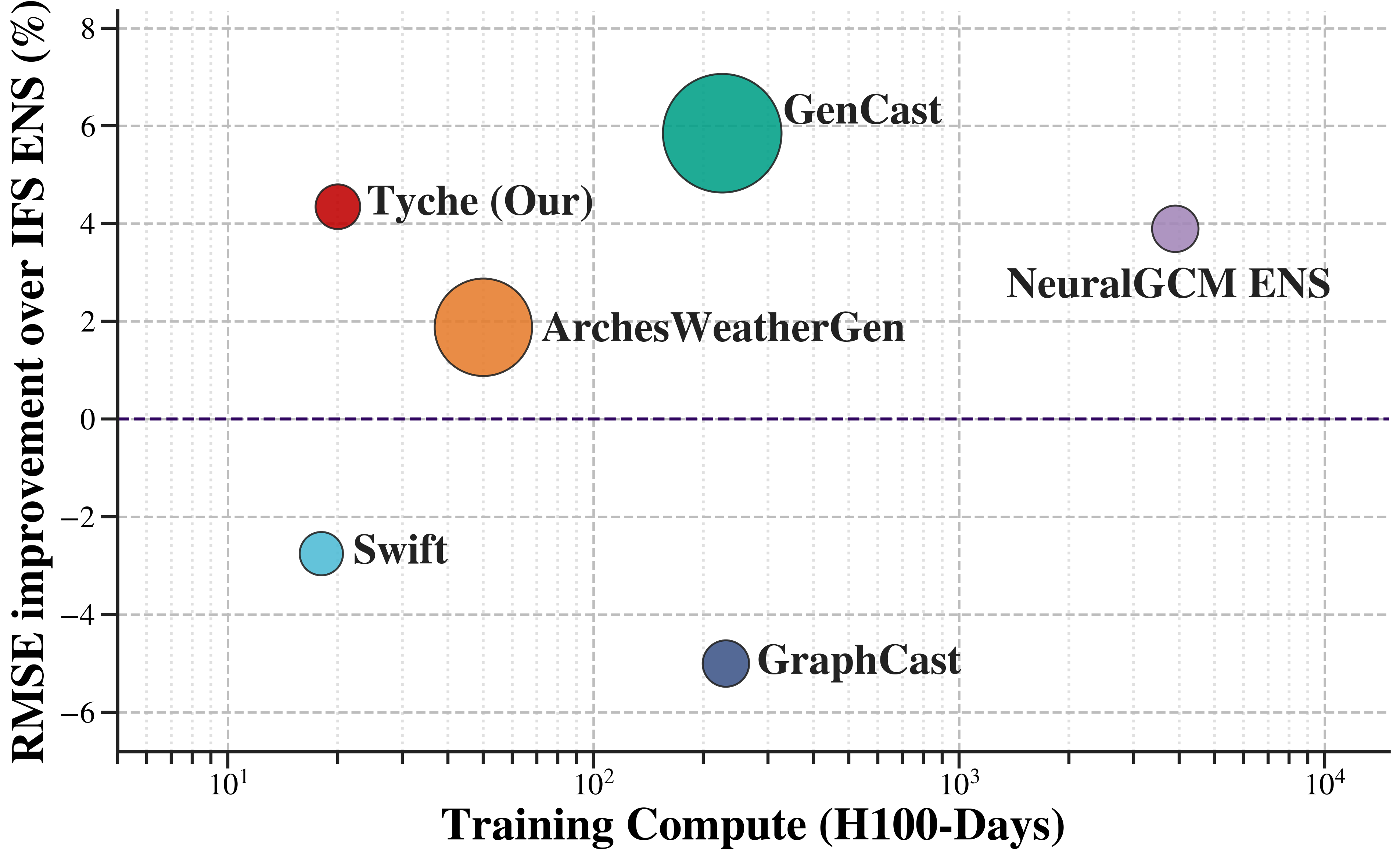}
    \caption{Visualization of forecast performance, training cost, and inference speed. The larger the circle, the slower the inference speed. }
    \label{fig:intro}
    \vspace{-20pt}
\end{wrapfigure}

While, one important remaining bottleneck is efficiency~\citep{utku2023efficient}. Existing probabilistic approaches typically require more than 20 function evaluations, resulting in over an order of magnitude higher computational cost than deterministic approaches. 
In global probabilistic forecasting, the cost of inference compounds along three axes: the rollout length \(L\), the ensemble size \(K\), and the number of denoising or integration steps \(N_{\mathrm{NFE}}\) required for each forecast transition. For diffusion and standard continuous-time flow models, this yields the unfavorable scaling $\text{total NFEs} \propto L \times K \times N_{\mathrm{NFE}}.$ When \(N_{\mathrm{NFE}} \gg 1\), the settings in which ensembles matter most, namely long forecast horizons and large ensemble sizes, are also the settings in which sampling becomes most expensive~\citep{andrae2025continuous,cachay2025elucidated,price2025probabilistic,zhuang2025ladcast}. Recent one-step models partially alleviate this tension~\citep{song2023consistency,lu2024simplifying,stock2025swift}, but its sampling methods are not flexible and often restricted to distillation frame. So the field still lacks a transport formulation explicitly designed for one-step conditional generation on high-dimensional geophysical states.

\textit{"For operational ensemble forecasting, the key design choice is not only the backbone, but the transport object."} Standard flow matching parameterizes an instantaneous velocity field and therefore inherits numerical integration at test time~\citep{lipman2022flow,wildberger2023flow,kornilov2024optimal}. By contrast, recent one-step generative work suggests that interval-averaged transport can make single-step sampling intrinsic rather than distilled~\citep{liu2022flow,geng2025mean}. We adopt this viewpoint for probabilistic weather forecasting and propose \ourmethod, a one-step conditional flow model targeting \emph{destination-aware average velocity}. Given a conditioning atmospheric state and a noisy latent state, \ourmethod{} predicts the average transport needed to move directly from the current point on the probability path to a chosen destination time, enabling direct \(1\)-NFE sampling.
The transport field is parameterized by an isotropic Swin-style transformer that preserves fixed spatial resolution, using local windows, shifted windows, and axial rotary position encoding to retain fine-scale meteorological structure on global grids. 
As efficient transport alone does not guarantee effective ensembles, we further separate transport learning from uncertainty calibration. Specifically, after one-step pretraining, we inference the model with parallel autoregressive ensemble rollouts, and finetune with a CRPS-based curriculum probabilistic loss.
As shown in Figure~\ref{fig:intro}, our method achieves highly competitive performance while maintaining low training and inference costs.

In summary, we make the following key contributions:
\ding{202} \textbf{\textit{Efficient Probabilistic Framework.}} 
We propose \ourmethod{}, a conditional 1-NFE flow-based generative framework specifically tailored for global probabilistic weather forecasting. \ourmethod{} effectively resolves the scalability bottleneck of multi-step diffusion models at the cost of deterministic baselines.
\ding{203} \textbf{\textit{Novel Two-Stage Strategy.}} 
To achieve distillation-free one-step generation on high dimensional atmospheric grids, we introduce a destination-aware average-velocity flow to bypass intractable computational overheads. Further, we propose a curriculum autoregressive finetuning strategy based on the CRPS objective to overcome the ensemble collapse and error accumulation issues.
\ding{204} \textbf{\textit{Superior Performance.}} 
Extensive experiments demonstrate that \ourmethod{} achieves highly calibrated ensembles and exceptional extreme weather prediction skills using merely a single function evaluation. To be specific, our proposed \ourmethod{} matches or exceeds the ensemble skill of state-of-the-art multi-step generative baselines, while running 12$-$43× faster than these generative baselines.

\section{Related Work}

\noindent\textbf{Probabilistic Weather Forecasting.}
Modern machine-learning weather prediction~\citep{chen2023fuxi,bodnar2025foundation,allen2025end,gao2025oneforecast} has been driven by highly accurate deterministic forecasters such as FourCastNet~\citep{pathak2022fourcastnet}, Pangu-Weather~\citep{bi2023accurate}, GraphCast~\citep{lam2023learning}, and NeuralGCM~\citep{kochkov2024neural}. While these models achieve strong medium-range skill, they do not explicitly model the conditional distribution of future states, and regression-based training tends to produce conditional-mean forecasts as uncertainty grows. To address this limitation, recent work has shifted toward probabilistic forecasting~\citep{li2025diffusion}, especially explicit generative modeling like diffusion. GenCast~\citep{price2025probabilistic} established diffusion as a strong paradigm for global probabilistic weather forecasting, showing that learned ensembles can achieve diverse samples and strong downstream utility. LaDCast~\citep{zhuang2025ladcast} moves diffusion into a compressed latent space to improve scalability. ERDM~\citep{cachay2025elucidated} models a forecast window jointly with progressively increasing uncertainty, adapting EDM design choices to sequence forecasting. However, despite their competitive performance, most existing methods remain constrained by multi-step denoising, with inference costs often exceeding those of deterministic models by more than an order of magnitude. In contrast, our method aims to achieve exceptional and efficient probabilistic forecasting through intrinsic one-step conditional transport. 
% slow sampling speed

\noindent\textbf{Efficient Generative Transport.}
Diffusion models and neural ODEs provide expressive continuous generative frameworks~\citep{chen2018neural,ho2020denoising,croitoru2023diffusion}, while flow matching formulates generation through learned instantaneous dynamics~\citep{lipman2022flow,liu2022flow,gat2024discrete}. However, these formulations typically require either multi-step denoising or numerical integration at inference time, which becomes increasingly expensive in high-dimensional autoregressive settings. Recent work therefore explores transport parameterizations that enable few-step or one-step generation, including consistency models~\citep{song2023consistency,lu2024simplifying}, Shortcut models~\citep{frans2024one}, and MeanFlow-style formulations based on interval-averaged transport~\citep{geng2025mean,geng2025improved,zhang2025alphaflow}. In weather forecasting, Swift~\citep{stock2025swift} is the closest 1-NFE counterpart, utilizing continuous-time consistency models to improve long-horizon ensemble forecasting efficiency. Instead of parameterizing instantaneous dynamics or a distilled consistency map, our method directly learns a destination-aware average velocity through JVP-regularized rectification for one-step conditional geophysical transport. 
% This makes one-step generation intrinsic to the model parameterization itself, while JVP-regularized rectification enforces temporal self-consistency and rollout-based CRPS finetuning improves ensemble reliability under autoregressive forecasting.

\section{Preliminaries}

\textbf{Problem setup.}
We consider probabilistic weather forecasting over a sequence of high-dimensional atmospheric fields $\{x^t\}_{t=1}^T$, where $x^t \in \mathbb{R}^{C \times H \times W}$ denotes the weather state at time $t$, with $C$ physical variables (\textit{e.g.}, \textit{temperature, geopotential, humidity}), and $H, W$ represent the spatial grid dimensions of latitude and longitude. Given the current observations $x^{t}$ as conditioning state, our goal is to learn the conditional probability distribution $p_{\psi}(x^{t+1} \mid x^{t})$ of the next future state $x^{t+1}$. Further, we model the forecast dynamics as a Markov chain, so that the conditional distribution at an arbitrary lead time $h$ can be obtained through recursive transitions. Following the standard autoregressive forecasting formulation, this conditional distribution can be factorized as:
\begin{equation}
    p_{\psi}\!\left(x^{t+1:t+h}\mid x^{t}\right) = \prod_{i=1}^{h} p_{\psi}\!\left(x^{t+i}\mid x^{t+i-1}\right).
\end{equation}
% \begin{equation}
%     p_{\theta}\!\left(x^{t+1:t+h}\mid x^{t-\ell+1:t}\right) = \prod_{i=1}^{h} p_{\theta}\!\left(x^{t+i}\mid x^{t-\ell+i:t+i-1}\right).
% \end{equation}
\textbf{Diffusion and Flow Matching formulation.}
Diffusion models~\citep{ho2020denoising,nichol2021improved} define a forward corruption process that gradually perturbs a clean target sample $x \sim p_{\mathrm{data}}(x)$ into noise over a continuous timestep $t \in [0,1]$. Specifically, the noisy state is constructed as $z_t = \varphi_t x + \sigma_t \epsilon$ with $\epsilon \sim \mathcal{N}(0,I)$, where $\varphi_t$ and $\sigma_t$ are pre-defined schedule coefficients that depend on $t$ and satisfy $z_0 = x$ and $z_1 = \epsilon$. Flow matching~\citep{lipman2022flow} provides a deterministic alternative by choosing a straight probability path between the noise distribution and the data distribution, setting $\varphi_t = 1 - t$ and $\sigma_t = t$:
\begin{equation}
    z_t = (1-t)x + t\epsilon, \quad v_t \triangleq v (z_t, t | x) = \epsilon - x,
\end{equation}
where $v_t$ denotes the corresponding ground-truth velocity along the path. A neural vector field \(v_{\theta}(z_t,t)\) is then trained to match this target velocity by minimizing:
\begin{equation}
    \mathcal{L}_{\mathrm{FM}}(\theta) = \mathbb{E}_{t,\,x,\,\epsilon} \left[ \left\| v_{\theta}(z_t,t) - v_t\right\|_2^2 \right].
\end{equation}
After training, generation is performed by solving the probability flow ODE (PF-ODE) $d z_t / dt = v_{\theta}(z_t,t)$ from \(t=1\) to \(t=0\), starting with an initial value \(z_1 \sim \mathcal{N}(0,I)\).

\section{Methodology}

\begin{figure*}[!t]
\centering
\includegraphics[width=1.0\linewidth]{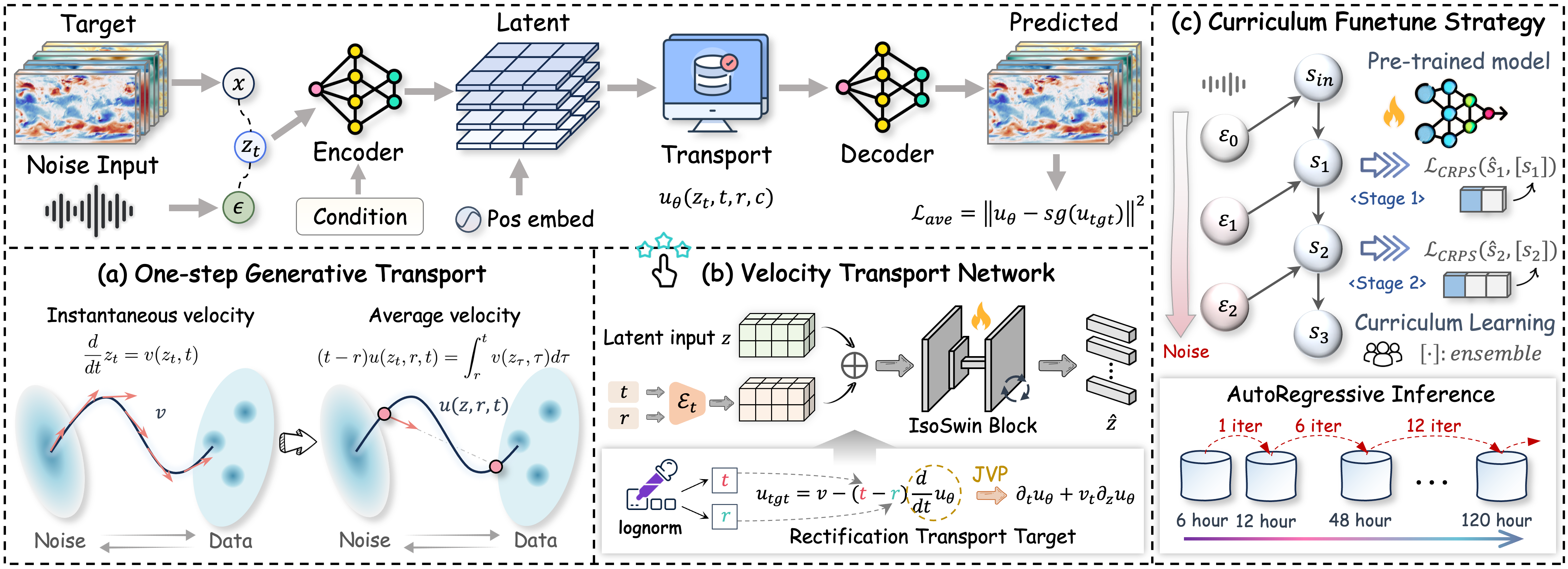}
\vspace{-10pt}
\caption{The overview of \ourmethod{}. 
\emph{(a):} Learn a average-velocity field with JVP-based rectification for direct one-step transport.
\emph{(b):} Parameterize the transport field with a conditional isotropic Swin-style backbone.
\emph{(c):} Finetune the transport model with curriculum autoregressive ensemble rollouts.
}
\label{fig:framework}
\vspace{-5pt}
\end{figure*}

\textbf{Framework Overview.}
Given a conditioning atmospheric state $c = x^{\tau}$, our goal is to model the conditional distribution of a future weather field $x = x^{\tau+1}$. 
As shown in Figure~\ref{fig:framework}, \ourmethod{} combines a destination-aware average-velocity flow formulation with a Swin-based velocity network. 
The flow formulation enables direct transport from Gaussian noise to a forecast state in one function evaluation, while the network provides a scalable parameterization for high-resolution global grids. 
Training is conducted in two stages: we first pretrain the transport field with a JVP-based rectification objective, and then finetune the model with rollout-based probabilistic supervision for ensemble calibration.

\subsection{One-Step Generative Transport via Rectified Average Velocity}

To distinguish the physical forecast horizon from the generative transport process, we reserve \(\tau\) for the physical forecast index and use \(t \in [0,1]\) to denote the continuous flow time. Conditioned on the current atmospheric state \(x^\tau\), we consider the standard linear probability path between the target future field \(x^{\tau+1}\) and Gaussian noise \(\epsilon \sim \mathcal{N}(0,I)\):
\begin{equation}
    z_t = (1-t)x^{\tau+1} + t\epsilon, \quad t \in [0,1].
    \label{eq:path}
\end{equation}
The reference instantaneous velocity driving this path is constant, namely \(v_t \triangleq \frac{\mathrm{d} z_t}{\mathrm{d} t} = \epsilon - x^{\tau+1}\).

Standard flow matching parameterizes an instantaneous vector field and typically requires multi-step numerical integration at inference time. Inspired by MeanFlow~\citep{geng2025mean}, we aim to directly learn the \emph{destination-conditioned average velocity} required to transport a state from the current flow time $t$ to a target destination time $r$ ($0 \le r < t \le 1$).
\begin{equation}
    u^\star(z_t,r,t,x^\tau)
    \triangleq
    \frac{1}{t-r}\int_r^t v(z_s,s,x^\tau)\,\mathrm{d}s = \frac{z_t-z_r}{t-r},
    \label{eq:avg_velocity}
\end{equation}
where \(v(z_s,s,x^\tau)\) denotes the underlying instantaneous velocity field of the conditional transport given \(x^\tau\). By definition, this average-velocity field induces the transport relation:
\begin{equation}
    z_r = z_t - (t-r)u^\star(z_t,r,t,x^\tau),
\end{equation}
Direct supervision of \(u^\star(z_t,r,t,x^\tau)\) is intractable because it requires evaluating the time integral in Eq.~\eqref{eq:avg_velocity}. To obtain a trainable objective, we exploit the differential identity satisfied by the average-velocity field (with respect to $t$, while $r$ is independent):
\begin{equation}
    u^\star(z_t,r,t,x^\tau) = v_t - (t-r) \frac{d}{dt} u^\star(z_t,r,t,x^\tau),
    \quad
    \frac{d}{dt} u^\star(z_t,r,t,x^\tau) \triangleq v_t\,\partial_z u^\star + \partial_t u^\star,
\end{equation}
where $u^\star$ is the abbreviation of $u^\star(z_t,r,t,x^\tau)$. We know that the total derivative can be interpreted as a Jacobian-vector product (JVP) between the Jacobian \([\partial_z u^\star, \partial_r u^\star, \partial_t u^\star]\) and the tangent vector \([v_t, 0, 1]\), and can therefore be computed efficiently using forward-mode automatic differentiation. We therefore parameterize a conditional transport field \(u^\star_\theta(z_t,r,t,x^\tau)\) and construct the practical rectified target:
\begin{equation}
    u_{\mathrm{tgt}}(z_t,r,t,x^\tau) = v_t - (t-r)\, (v_t\,\partial_z u_\theta^\star + \partial_t u_\theta^\star)).
    \label{eq:utgt}
\end{equation}
We then optimize the first stage transport objective:
\begin{equation}
    \mathcal{L}_{\mathrm{ave}}(\theta) =
    \mathbb{E}_{\epsilon,\, x^\tau,\,x^{\tau+1}}
    \,
    \mathbb{E}_{t \sim \mathcal{U}[0,1],\, r \sim \mathcal{U}[0,t]}
    \left[
        \left\|
        u_\theta(z_t,r,t,x^\tau) - \operatorname{sg}( u_{\mathrm{tgt}}(z_t,r,t,x^\tau) )
        \right\|_F^2
    \right],
    \label{eq:lflow}
\end{equation}
where \(\operatorname{sg}(\cdot)\) denotes the stop-gradient operator. At inference time, one-step conditional generation is obtained by evaluating the learned average-velocity field only once at the noise endpoint:
\begin{equation}
    \hat{x}^{\tau+1}
    =
    \epsilon - u_\theta(\epsilon,0,1,x^\tau),
    \label{eq:one_step_sampling}
\end{equation}
which produces a forecast sample with a single neural function evaluation (1-NFE).

\subsection{Velocity Transport Network Parameterization}
\textbf{Temporal Conditioning and Adaptive Modulation.}
We parameterize the average-velocity field \(u_\theta(z_t,r,t,x^{\tau})\) with a conditional isotropic Swin-style diffusion transformer. We first concatenate the noisy state $z_t$ and the conditioning field $x^{\tau}$, patchify the corresponding tensor, and add learned absolute positional embeddings $P_{abs}$ to encode fixed geographic structure. 
Further, we sample the source and destination flow timesteps $(r,t) \in [0,1]$ from a logit-normal distribution. By the way, for a fixed fraction of samples, we set $r=t$ to expose the model to the boundary case that recovers standard flow matching. 
To condition the transport field on both the current flow time \(t\) and the destination time \(r\), we encode them separately with sinusoidal timestep embeddings followed by two independent MLPs, and sum the resulting embeddings. This temporal representation modulates each transformer block through Adaptive RMSNorm~\citep{jiang2023pre}:
\begin{equation}
\begin{aligned}
    t_{\mathrm{emb}} = f_t(\phi(t)) + & f_r(\phi(r)), \quad
    [\alpha^{(l)},\beta^{(l)},\gamma^{(l)}] = 
    W_{\mathrm{mod}}^{(l)} t_{\mathrm{emb}} + b_{\mathrm{mod}}^{(l)}, \\
    \mathcal{M}^{(l)}(z) &= \operatorname{RMSNorm}(z) \odot
    \bigl(1+\alpha^{(l)}\bigr) + \beta^{(l)}.
\end{aligned}
\end{equation}
Here \(\boldsymbol{\gamma}^{(l)}\) acts as a channel-wise residual gate. The modulation layers are zero-initialized, making each residual branch identity-like at initialization and stabilizing early-stage training.

\textbf{Isotropic Swin Transformer.}
Following the Swin Transformer paradigm, self-attention is constrained within local windows of size $W_h \times W_w$, reducing the computational complexity from $\mathcal{O}(N^2)$ to $\mathcal{O}(N \cdot W_h W_w)$. To capture the relative movements of advective flows and vortices, we apply 2D Axial Rotary Positional Embeddings (RoPE)~\citep{heo2024rotary} to the queries \textit{Q} and keys \textit{K} along both $X$ and $Y$ spatial axes inside the local window prior to the dot-product attention calculation.

To reconstruct global atmospheric teleconnections without destroying the isotropic resolution, we alternate between regular and shifted window partitioning across consecutive layers. Let $\mathcal{S}$ denote the cyclic shift operator applied at odd layers to cross window boundaries, yielding the pre-aligned feature $\hat{z}^{(l)} = \mathcal{S}(z^{(l-1)})$. Integrating these mechanisms, the full forward pass for the $l$-th \textit{IsoSwin Block} is formulated as a gated dynamical system:
\begin{equation}
\begin{aligned}
    \tilde{z}^{(l)} &= \hat{z}^{(l)} + \boldsymbol{\gamma}_{\mathrm{att}}^{(l)} \odot \operatorname{WAttn}_{\mathrm{RoPE}}\left( \mathcal{M}_{\mathrm{att}}^{(l)}(\hat{z}^{(l)}) \right), \\
    z^{(l)} &= \mathcal{S}^{-1} \Big( \tilde{z}^{(l)} + \boldsymbol{\gamma}_{\mathrm{ffn}}^{(l)} \odot \operatorname{SwiGLU}\left( \mathcal{M}_{\mathrm{ffn}}^{(l)}(\tilde{z}^{(l)}) \right) \Big),
    \label{eq:full_block}
\end{aligned}
\end{equation}
where $\mathcal{S}^{-1}$ reverses the cyclic shift to align the spatial grid for the next layer, and SwiGLU~\citep{shazeer2020glu} acts as the channel-mixing feed-forward network. After the final block, a lightweight output head linearly maps the latent tokens back to patch values and unpatchifies them to the physical grid, yielding the predicted transport field \(u_\theta(z_t,r,t,x^{\tau})\). 
% Detailed layer-wise formulations are illustrated in Appendix.
Detailed layer-wise formulations and implementation specifics are deferred to the Appendix~\ref{app:network}.

\subsection{Two-Stage Training Objective and Ensemble Inference}

$\star$ \textit{\textbf{Stage I: One-step transport pretraining.}}
We first pretrain the transport network to learn a one-step average-velocity field. For each training pair \((x^\tau,x^{\tau+1})\), we sample Gaussian noise \(\epsilon\sim\mathcal{N}(0,I)\) and flow-time pair $r,t \in [0,1]$. Then we utilize the $\mathcal{L}_{\mathrm{ave}}(\theta)$ and the corresponding Jacobian-vector product to supervise the direct average-velocity transport update $u_\theta(\epsilon,0,1,x^\tau)$.

$\star$ \textit{\textbf{Stage II: Curriculum probabilistic calibration.}}
Although Stage I learns accurate local transport, recursive weather forecasting additionally requires calibrated uncertainty under autoregressive rollout. We therefore finetune the pretrained model using short ensemble rollouts and gradually increase the rollout horizon during training. At curriculum stage \(m\), let \(R_m\) be the current rollout length. Starting from \(c_0=x^\tau\), each ensemble member is generated recursively by:
\begin{equation}
    \hat{x}^{\tau+l}_{k} = \epsilon^{l}_{k} -
    u_\theta\!\left( \epsilon^{l}_{k},0,1,c^{l-1}_{k} \right),
    \quad
    \epsilon^{l}_{k}\sim\mathcal{N}(0,I),
    \quad
    k=1,\dots,K ,
\end{equation}
where \(c^{l-1}_{k}\) is the conditioning state formed from the previous prediction, and $K$ represents the number of ensemble members. Given the ensemble \(\{\hat{x}^{\tau+l}_{k}\}_{k=1}^{K}\) and the ground-truth state \(x^{\tau+l}\) at lead time $l$, we optimize the empirical Continuous Ranked Probability Score (CRPS):
\begin{equation}
\mathcal{L}_{\mathrm{CRPS}}(l) = 
    % \operatorname{CRPS}_{K}
    % \!\left(
    % \{\hat{x}^{\tau+l}_{k}\}_{k=1}^{K},
    % x^{\tau+l}
    % \right)
    % =
    \frac{1}{K}\sum_{k=1}^{K}
    \left\|
    \hat{x}^{\tau+l}_{k}-x^{\tau+l}
    \right\|_1
    -
    \frac{1}{2K(K-1)}
    \sum_{k=1}^{K}\sum_{k'=1}^{K}
    \left\|
    \hat{x}^{\tau+l}_{k}
    -
    \hat{x}^{\tau+l}_{k'}
    \right\|_1 .
    \label{eq:empirical_crps}
\end{equation}
The first term encourages ensemble accuracy, while the second term discourages ensemble collapse by rewarding appropriate dispersion. Rather than supervising every intermediate rollout step, we evaluate the calibration loss only on the terminal forecast state of the rollout. In our experimental setting, we do not perform many rounds of iterative finetuning. We set \(m=2\), with the corresponding rollout horizons \(R_m\) set to \(1\) and \(2\), respectively. Further details are provided in the experiments.

$\star$ \textit{\textbf{1-NFE Ensemble Inference.}}
To generate a $K$-member forecast ensemble, we sample $K$ independent noise instances, set the current state to pure noise ($t=1$), and the destination to the target forecast ($r=0$). The future ensemble is generated via a single function evaluation (1-NFE):
\begin{equation}
    \hat{x}^{\tau+l}_{k} = \epsilon^{l}_{k} - u_\theta\!\left(
    \epsilon^{l}_{k},0,1,x^{\tau+l-1}_{k} \right),
    \quad l=1,\dots,H,\quad k=1,\dots,K .
    \label{eq:ensemble_inference}
\end{equation}
Naturally, all \(K\) ensemble members for the next time step can be generated in parallel on a single GPU with a single batched inference pass.
This circumvents the computationally prohibitive multi-step ODE solvers, reducing the generation time of high-fidelity weather ensembles from minutes to milliseconds, while maintaining physically consistent spatial spread.

\section{Experiment}
\label{experiment}

\subsection{Experimental Settings}
\label{exp:setting}

\textbf{Datasets.} 
We train and evaluate \ourmethod on the ERA5 reanalysis dataset~\cite{hersbach2020era5} at 1.5° resolution (120 $\times$ 240 grid) and 6-hourly interval. \ourmethod is trained to forecast nine variables: four surface-level variable (\texttt{u10}, \texttt{v10}, \texttt{t2m}, \texttt{mslp}) and five atmospheric variables (\texttt{z}, \texttt{q}, \texttt{t}, \texttt{u}, \texttt{v}) at 13 pressure levels (\texttt{50}, \texttt{100}, \texttt{150}, \texttt{200}, \texttt{250}, \texttt{300}, \texttt{400}, \texttt{500}, \texttt{600}, \texttt{700}, \texttt{850}, \texttt{925}, \texttt{1000}). Totally, the data contains 69 channels. Before training, all variables are standardized by subtracting their respective means and dividing by their standard deviations. Specifically, we utilize data from 1979–2017 for training, 2018-2019 for validation, and 2020 for testing. More details are illustrated in Appendix~\ref{app:data}.

% \ding{253} \ding{95} 
\textbf{Baselines.}
To comprehensively evaluate \ourmethod, we conduct experiments against two main categories of state-of-the-art baselines: deterministic approaches and probabilistic approaches. \ding{69} For deterministic baselines, we choose Pangu~\cite{bi2023accurate} and GraphCast~\cite{lam2023learning}. These two are based on advanced Transformer or GNN-style architectures. \ding{69} For probabilistic baselines, we select ARCI~\cite{andrae2025continuous}, Swift~\cite{stock2025swift}, ArchesWeatherGen (ArchesGen)~\cite{couairon2024archesweather}, GenCast~\cite{price2025probabilistic}, and NeuralGCM-ENS~\cite{kochkov2024neural}. All these approaches are combination of neural networks and Diffusion models. Fur better evaluate, we also compare with ECMWF IFS ensemble (IFS-ENS)~\cite{rasp2024weatherbench}. See Appendix~\ref{app:baseline} for detailed descriptions.

\textbf{Implementation details.} 
For \ourmethod, training pipeline is divided into two stages. In Stage I, the model is trained using the MSE loss over 300 epochs. We used the AdamW optimizer with an initial learning rate of $10^{-4}$, scheduled by a cosine annealing policy with a minimum learning rate of $10^{-6}$. In Stage II, we perform curriculum autoregressive training using the empirical CRPS loss with an ensemble size of 2. The rollout horizon progressively increases from 1 to 2. The learning rate is reduced to $10^{-5}$ for this stage. The architecture backbone consists of 12 transformer blocks, configured with a hidden dimension of 1024 and 12 attention heads. Spatial interactions are restricted to local windows of size $10 \times 10$, which are cyclically shifted by $5 \times 5$ in alternating blocks. In our setting, it has around 230 MB trainable parameters.
For compared baselines except GenCast, NeuralGCM-ens, and IFS-ENS, we utilize the official implementation for all models and train from scratch to ensure relative fairness. Specifically, for other three, we download the saved inference data of corresponding year with certain dimensions from Weatherbench2. The ensemble number is as the same of other baselines. 
To ensure computational efficiency, we conducted all training on eight 80GB NVIDIA H100 GPUs, and inference on a single H100 GPU. More details see in Apendix~\ref{app:baseline}.

\textbf{Evaluation.} 
We aim to generate 20 members for each compared model. Specifically, for deterministic approaches, we add independent noise to initial state; and for probabilistic approaches, we directly generate 20 members parallelly. For evaluation, we report metrics like latitude-weighted root mean square error (RMSE) and continuous ranked probability score (CRPS). Also, we report spread to skill ratio (SSR) to quantify the uncertainty capability. More details are demonstrated in Appendix~\ref{app:metrics}.

\begin{table*}[t]
\centering
% \small
\setlength{\tabcolsep}{4pt}  % 每一列左右两侧的内边距
\renewcommand{\arraystretch}{1.2}  % 行距倍数设为 1.3
\caption{We evaluate the performance of our \ourmethod again five baselines (two deterministic baselines and three probabilistic baselines) up to 10 days (40 iteration). We report the RMSE results of three variables: \texttt{u10}, \texttt{t2m}, and \texttt{mslp}, and a small RMSE ($\downarrow$) identify better performance.} 
% \vspace{-5pt}
\label{tab:main_results}
\begin{sc}
\resizebox{1.0\textwidth}{!}{
\begin{tabular}{lccccccccccccccc}
\toprule
\multirow{2}{*}{Model}
& \multicolumn{3}{c}{6-Hour}
& \multicolumn{3}{c}{1-Day}
& \multicolumn{3}{c}{3-Day}
& \multicolumn{3}{c}{7-Day}
& \multicolumn{3}{c}{10-Day}\\
\cmidrule(lr){2-4} \cmidrule(lr){5-7} \cmidrule(lr){8-10} \cmidrule(lr){11-13} \cmidrule(lr){14-16}
& \texttt{u10} & \texttt{t2m} & \texttt{mslp} & \texttt{u10} & \texttt{t2m} & \texttt{mslp} & \texttt{u10} & \texttt{t2m} & \texttt{mslp} & \texttt{u10} & \texttt{t2m} & \texttt{mslp} & \texttt{u10} & \texttt{t2m} & \texttt{mslp} \\
\midrule
Pangu & 0.58 & 0.92 & 0.64 & 1.14 & 1.50 & 1.47 & 2.53 & 1.90 & 3.62 & 4.05 & \underline{\second{3.23}} & 7.02 & \underline{\second{4.85}} & \underline{\second{4.05}} & 10.29 \\
GraphCast & \best{0.48} & \best{0.75} & \underline{\second{0.53}} & \best{0.96} & \underline{\second{1.24}} & \underline{\second{1.09}} & 2.21 & 2.28 & 2.86 & \underline{\second{3.97}} & 3.62 & \underline{\second{6.64}} & 4.94 & 4.50 & 10.28 \\
% \midrule
ARCI & 0.89 & 1.41 & 0.95 & 1.45 & 2.40 & 1.63 & 2.75 & 3.55 & 3.50 & 5.32 & 5.23 & 8.47 & 5.92 & 6.37 & 11.26 \\
Swift & 0.85 & 1.32 & 0.88 & 1.46 & 2.32 & 1.56 & 2.81 & 3.32 & 3.26 & 5.14 & 5.11 & 7.54 & 5.24 & 5.58 & 10.49 \\
ArchesGen & 0.62 & 0.82 & 0.57 & 1.11 & 1.25 & 1.27 & \underline{\second{2.15}} & \underline{\second{1.88}} & \underline{\second{2.71}} & 4.25 & 3.35 & 6.84 & 4.88 & 4.17 & \underline{\second{9.91}} \\
Tyche (Our) & \underline{\second{0.51}} & \underline{\second{0.76}} & \best{0.49} & \underline{\second{1.09}} & \best{1.19} & \best{0.93} & \best{2.13} & \best{1.68} & \best{2.49} & \best{3.89} & \best{2.77} & \best{6.03} & \best{4.71} & \best{3.74} & \best{8.58} \\
\bottomrule
\end{tabular}
}
\end{sc}
\vspace{-5pt}
\end{table*}

\begin{figure*}[!t]
\centering
\includegraphics[width=1.0\linewidth]{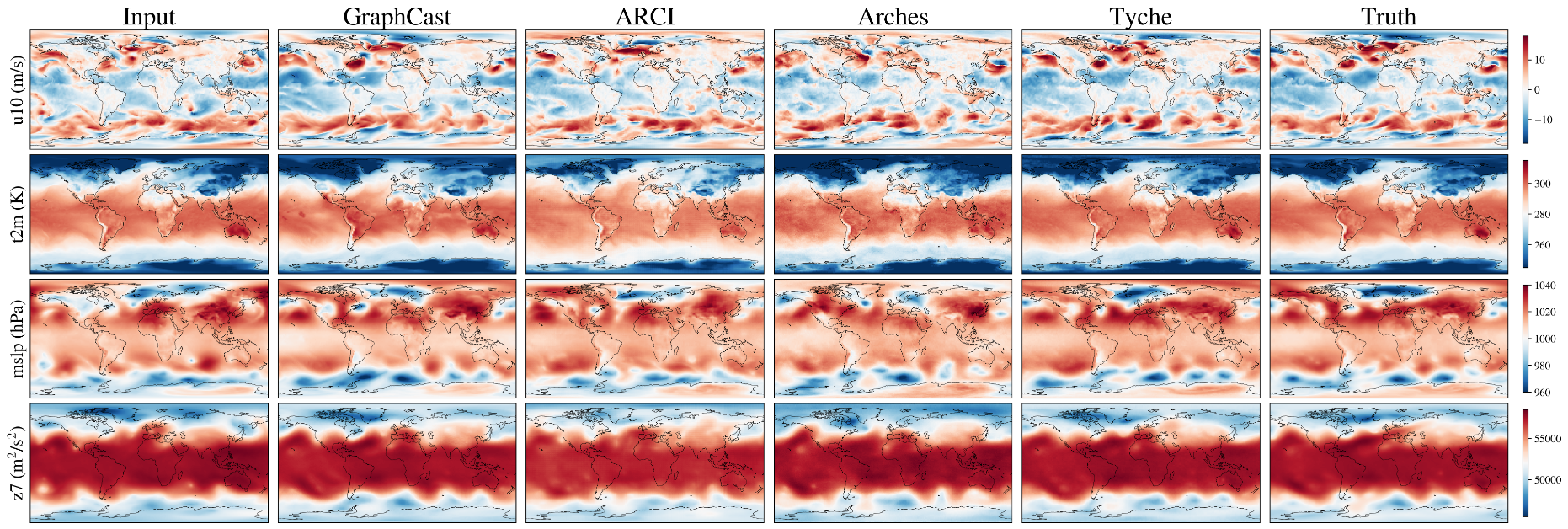}
\vspace{-15pt}
\caption{Comparison of 7-day forecasts for four variables: \texttt{u10}, \texttt{t2m}, \texttt{mslp}, and \texttt{z7}. 
}
\label{fig:exp1}
\vspace{-5pt}
\end{figure*}

\subsection{Main Results}

\textbf{Forecasting Performance.}
To evaluate the effectiveness of our method, we first compare \ourmethod with two deterministic baselines and three probabilistic baselines. Both of them are trained from scratch. Regarding the number of sampling steps during inference, we use corresponding multi-steps sampling for ARCI and ArchesGen. While for Swift and our method, which are designed for one-step generation, we use single-step sampling. 
Here, we do not use the ensemble mean of probabilistic models for comparison, we instead select only one inference member. Table~\ref{tab:main_results} reports the latitude-weighted RMSE of \ourmethod and five baselines from 6 hours to 10 days on three representative surface variables, namely \texttt{u10}, \texttt{t2m}, and \texttt{mslp}. Overall, \ourmethod consistently achieves the best or second-best performance across almost all variables and forecast horizons. We observe that within the first day of forecasting, GraphCast achieves the best performance across multiple variables, while Pangu also delivers competitive results. This suggests that deterministic models retain a clear advantage at short lead times, where error accumulation remains limited. At 3 days, \ourmethod achieves the lowest RMSE on all three variables. Compared with GraphCast, \ourmethod reduces the 3-day RMSE from 2.21 to 2.13 on \texttt{u10}, from 2.28 to 1.68 on \texttt{t2m}, and from 2.86 to 2.49 on \texttt{mslp}. Typically, when the forecast horizon extending 3-day, \ourmethod still maintains the best performance across all three variables. 

Figure~\ref{fig:exp1} provides a qualitative comparison of 7-day forecasts for \texttt{u10}, \texttt{t2m}, \texttt{mslp}, and \texttt{z7}, where \texttt{z7} represents geopotential at 500 hPa. The visual results are consistent with the quantitative findings in Table~\ref{tab:main_results}. Compared with other baselines, \ourmethod better preserves large-scale spatial structures and produces forecast fields that are visually closer to the ground truth. For \texttt{u10}, \ourmethod captures the major wind patterns in the mid-latitudes and Southern Ocean with less over-smoothing. For \texttt{t2m}, it maintains coherent meridional temperature gradients and more realistic regional variations. For \texttt{mslp}, \ourmethod better reconstructs large-scale pressure systems. For \texttt{z7}, \ourmethod also preserves the global geopotential structure more faithfully, indicating improved stability for upper-air variables. In summary, the quantitative and qualitative results jointly demonstrate the effectiveness of \ourmethod.

\begin{figure*}[!t]
\centering
\includegraphics[width=1.0\linewidth]{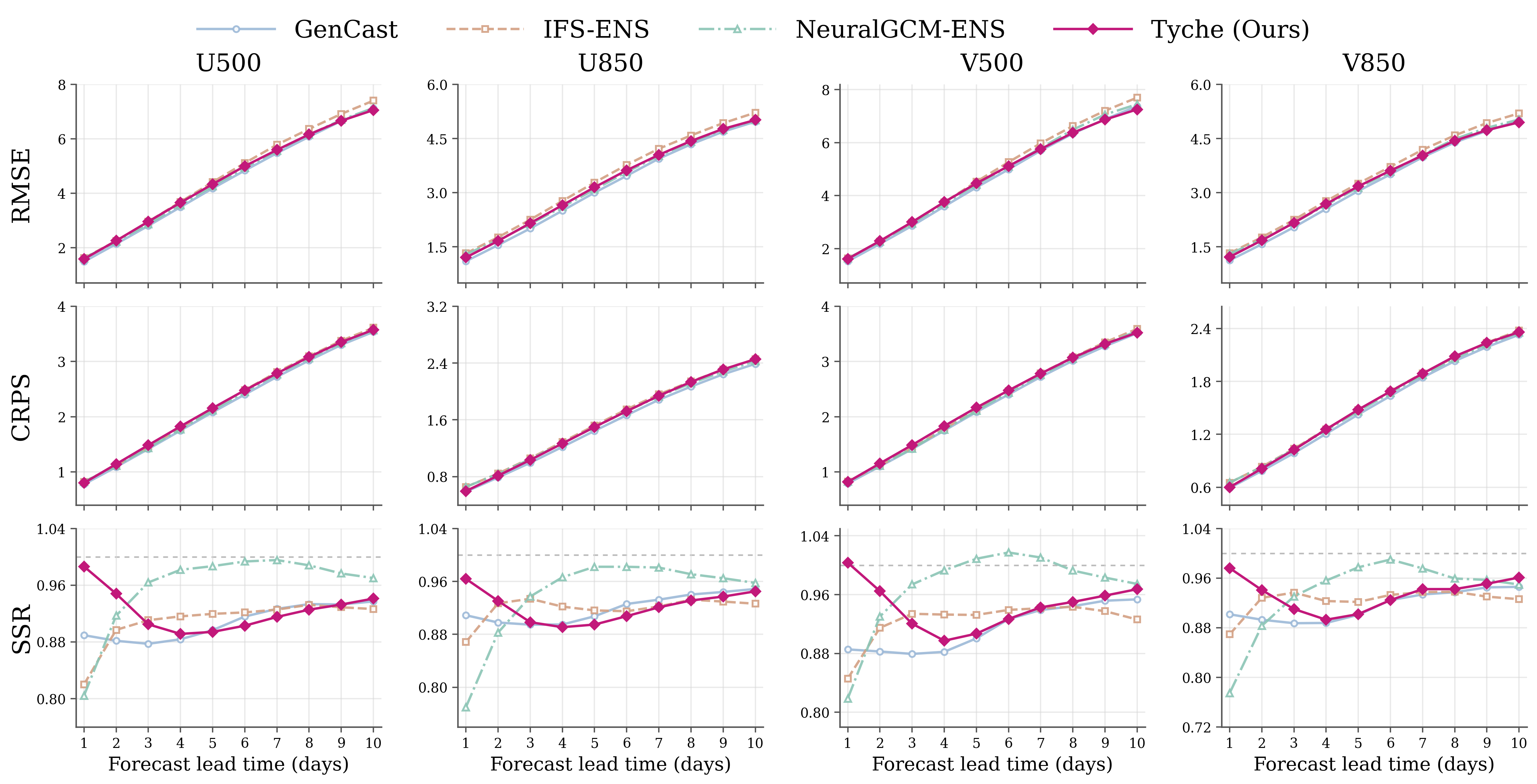}
\vspace{-5pt}
\caption{Ensemble Skill Curves Across Forecast Lead Time. We report the latitude-weighted RMSE ($\downarrow$), latitude-weighted CRPS ($\downarrow$), and SSR ($\to 1$) of several upper-air variables.
}
\label{fig:exp2}
\vspace{-5pt}
\end{figure*}

\begin{figure*}[!t]
\centering
\includegraphics[width=1.0\linewidth]{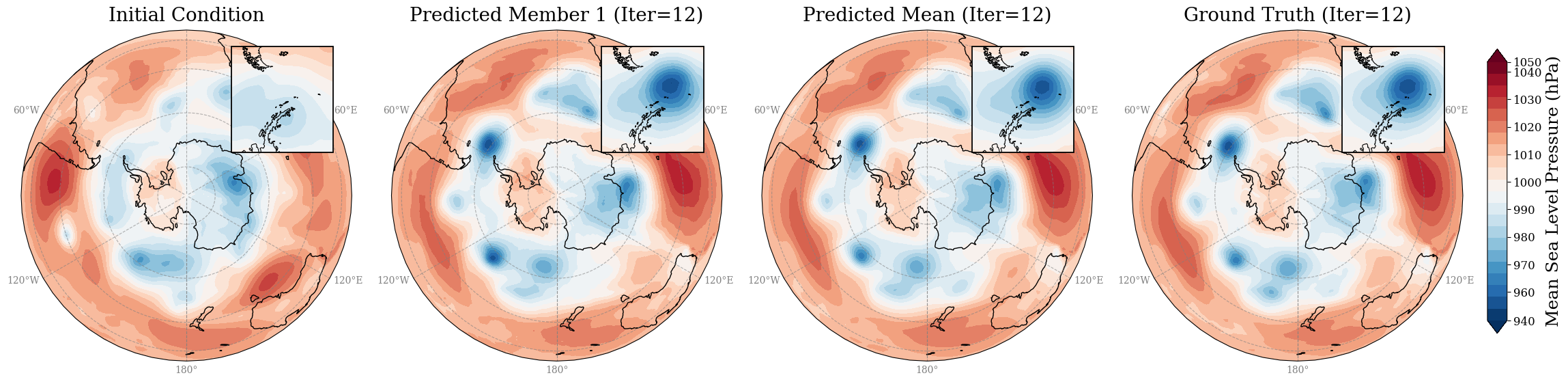}
\vspace{-5pt}
\caption{Visualization of 3-day MSLP ensemble forecast performance over the Southern Ocean. 
}
\label{fig:exp4}
\vspace{-12pt}
\end{figure*}

\textbf{Ensemble Performance.}
Further, we evaluate the probabilistic forecasting skills of \ourmethod by comparing it with state-of-the-art ensemble forecasting systems, including GenCast, NeuralGCM-ENS, and IFS-ENS. To ensure a fair comparison, we use 20 ensemble members for both \ourmethod and all compared ensemble baselines. Figure~\ref{fig:exp2} reports the latitude-weighted RMSE, CRPS, and SSR over a 10-day forecast horizon on key upper-air variables, including \texttt{U500}, \texttt{U850}, \texttt{V500}, and \texttt{V850}. Unlike the single-member comparison in Table~\ref{tab:main_results}, this evaluation focuses on the performance of the ensemble skills. 
As illustrated in Figure~\ref{fig:exp2}, the curves of \ourmethod follow a similar growth trend to established ensemble baselines, indicating that the proposed one-step probabilistic generator does not suffer from rapid error divergence during autoregressive rollout. 
Specifically, \ourmethod demonstrates competitive performance against compared systems in terms of RMSE. Meanwhile, the CRPS curves show that \ourmethod produces meaningful ensemble distributions rather than merely accurate individual samples, demonstrating its ability to capture forecast uncertainty over medium-range lead times. Additionally, the SSR curves rigorously validate the model's probabilistic calibration. Following an initial short-term adjustment, the SSR of \ourmethod converges tightly toward the optimal value of 1. 
As known, GenCast follows a conventional diffusion-based generative paradigm and requires more than 20 denoising steps during inference, while NeuralGCM-ENS incorporates physics-based numerical solvers into the forecasting process. Meanwhile, \ourmethod is a purely data-driven probabilistic model and performs only single-step sampling at inference time. 
Beyond aggregated metrics, \ourmethod demonstrates exceptional physical fidelity in capturing high-frequency, synoptic-scale structures. As visualized in Figure~\ref{fig:exp4}, we examine the Mean Sea Level Pressure (\textit{MSLP}) over the dynamically volatile Southern Ocean at a 3-day lead time (\textit{Iteration=12}). An individual forecast (\textit{Predicted Member 1}) exhibits physically realistic vortex structures that strictly adhere to the ground truth. And the predicted mean displays the characteristic smoothing expected from a well-calibrated ensemble, without the appearance of blur. 
This demonstrates the superior ensemble performance of our method.

\begin{wraptable}{r}{0.65\textwidth}
% \vspace{-10pt}
\renewcommand{\arraystretch}{1.3}
  \centering
  \caption{Computational efficiency and inference comparison. }
  \vspace{-5pt}
  \label{tab:efficiency_comparison}
  % \begin{sc}
  \resizebox{\linewidth}{!}{%
    \begin{tabular}{lccc}
        \toprule
        \textbf{Model} & \textbf{Methodology} & \textbf{NFE per Rollout} & \textbf{Inference Latency} \\
        \midrule
        IFS-ENS & Operational NWP & N/A (Physics)  & Hours \\
        \midrule
        GenCast & EDM & $20$  & $\sim$ 45 mins \\
        ARCI & EDM & $20$   & $\sim$ 12 mins \\
        ArchesWeatherGen & Flow Matching & $25$   & $\sim$ 36 mins \\
        \rowcolor{gray!10} \textbf{Tyche (Our)} & \textbf{Average Transport} & $\mathbf{1}$   & \textbf{$\sim$ 62 secs} \\
        \bottomrule
    \end{tabular}
  }
  % \end{sc}
\vspace{-10pt}
\end{wraptable}

\textbf{Efficiency Analysis.}
The primary bottleneck of modern probabilistic weather forecasting lies in the inference computational cost, which strictly compounds along four axes: the forecast rollout length ($L$); the ensemble size ($K$); the number of function evaluations ($N_{\text{NFE}}$) per rollout; and the inference cost ($M$) per NFE. Consequently, the total inference cost scales as $\mathcal{O}(L \times K \times N_{\text{NFE}} \times M)$. Multi-step diffusion models (\textit{e.g.}, GenCast, ArchesWeatherGen) typically require $N_{\text{NFE}} \ge 10$ per rollout, making the generation of multi-day high-fidelity weather ensembles take minutes and rendering large-scale operational deployment prohibitively expensive. \ourmethod fundamentally breaks this computational bottleneck through its average velocity transport formulation. 
% In this way, \ourmethod reduces the evaluation cost of a single ensemble member to the similar time complexity as deterministic methods. 
As detailed in Table \ref{tab:efficiency_comparison}, \ourmethod can parallelly generate a 20-member, 15-day trajectory (60 autoregressive iteration) in merely $\sim$ 62 seconds. This achieves an order-of-magnitude acceleration, running approximately 12× to 43× faster than current probabilistic MLWP systems (with the same resolution of data). 
Thus, \ourmethod matches the inference speeds of deterministic methods, while preserving the robust diversity and calibration typically restricted to costly multi-step diffusion and operational NWP systems.

\begin{wrapfigure}[11]{r}{0.65\linewidth}
    \centering
    \vspace{-10pt}
    \includegraphics[width=1.0\linewidth]{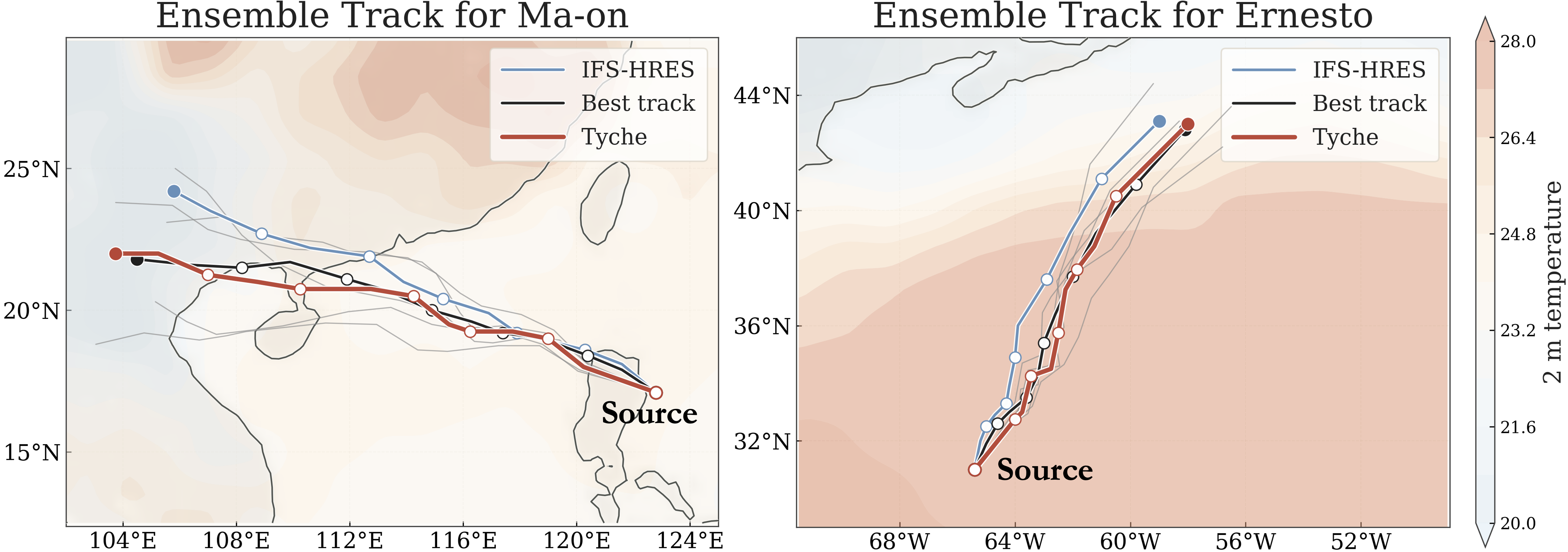}
    \caption{Typhoon ensemble track for Ma-on and Ernesto.}
    \label{fig:typhoon}
    \vspace{-5pt}
\end{wrapfigure}

\textbf{Typhoon Tracking.}
We further evaluate \ourmethod's capability in predicting high-impact synoptic-scale events through tropical cyclone tracking. Figure~\ref{fig:typhoon} compares the trajectories for Typhoon Ma-on \texttt{(2022/08)} and Ernesto \texttt{(2024/08)}. Crucially, the individual ensemble members of \ourmethod (thin grey lines) form a well-calibrated cone of uncertainty. And the ensemble average trajectory fits the best trajectory very well. For the westward-moving Ma-on, the ensemble mean of \ourmethod (thick red line) tightly adheres to the ground truth, accurately reproducing the landfall direction. For the northeastward-curving Ernesto, \ourmethod successfully captures the complex recurvature dynamics, aligning perfectly with the late-stage evolution where IFS-HRES exhibits a noticeable westward cross-track error. These findings suggest that the proposed one-step transport effectively preserves coherent vortex motion, offering robust and scalable probabilistic forecasts for extreme weather phenomena.

\begin{wraptable}{r}{0.65\textwidth}
\vspace{-10pt}
\renewcommand{\arraystretch}{1.3}
  \centering
  \caption{Ablation study on generative transport strategy. }
  \vspace{-5pt}
  \label{tab:ablation}
  \begin{sc}
  \resizebox{\linewidth}{!}{%
    \begin{tabular}{c|cc|cc|cc}
    \toprule
    \multirow{2}{*}{Method} 
    & \multicolumn{2}{c|}{1-NFE} 
    & \multicolumn{2}{c|}{3-NFE} 
    & \multicolumn{2}{c}{10-NFE} \\
    \cmidrule(lr){2-3}
    \cmidrule(lr){4-5}
    \cmidrule(lr){6-7}
    & Step-1 & Step-10
    & Step-1 & Step-10
    & Step-1 & Step-10 \\
    \midrule
    w/ EDM      
    & -- & -- & 0.145 & 0.428 & 0.108 & 0.245 \\
    w/ TrigFlow 
    & -- & -- & 0.134 & 0.390 & \underline{0.101} & 0.250 \\
    w/ sCM      
    & 0.154 & 0.462 & 0.151 & 0.454 & -- & -- \\
    Ours 
    & 0.103 & \underline{0.242} & \textbf{0.099} & \textbf{0.237} & -- & -- \\
    \bottomrule
    \end{tabular}
  }
  \end{sc}
\vspace{-10pt}
\end{wraptable}

\textbf{Ablation on Generative Transport Strategy.}
We further investigate the effectiveness of generative transport strategy. Specifically, we conduct an ablation study comparing \ourmethod against three alternative state-of-the-art generative transport strategies: \ding{68} \textit{\textbf{EDM}} (\textit{a unified diffusion model})~\citep{karras2022elucidating}, \ding{68} \textit{\textbf{Triglow}} (\textit{a customized v-prediction flow matching})~\citep{lu2024simplifying}, and \ding{68} \textit{\textbf{sCM}} (\textit{the continuous-time consistency model})~\citep{lu2024simplifying}. To isolate the effect of the generative transport strategy, we conduct all ablation experiments under a unified model architecture and vary only the diffusion or flow training strategy. For each compared strategy, we follow the hyperparameter settings recommended in the original paper. We evaluate each method under three sampling budgets, namely 1-NFE, 3-NFE, and 10-NFE, and report RMSE within two autoregressive horizons (the 1-step and 10-step rollouts mean), as shown in Table~\ref{tab:ablation}.
The results reveal the efficiency tradeoff for conventional multi-step generative samplers. Both TrigFlow and EDM benefit from increasing the number of function evaluations. 
% TrigFlow improves from 0.134/0.390 at 3-NFE to 0.097/0.250 at 10-NFE, while EDM improves from 0.145/0.428 to 0.103/0.242. 
By contrast, sCM supports fast one-step generation, but its rollout error remains high. In contrast, \ourmethod achieves strong performance with only a single function evaluation. At 1-NFE, \ourmethod obtains 0.103 RMSE at Step-1 and 0.242 RMSE at Step-10, matching the 10-NFE EDM result while using only one tenth of the sampling budget. With 3-NFE inference, \ourmethod further improves the Step-1 and Step-10 RMSE to 0.099 and 0.237, achieving the best performance among all compared settings.

\section{Conclusion \& Limitation}
In this paper, we introduced Tyche, a fundamentally new generative framework that breaks the computational bottleneck of multi-step diffusion and flow models in global probabilistic weather forecasting. Tyche achieves 1-NFE generation by learning a destination-aware average-velocity flow. Further, we effectively mitigated the theoretically bounded autoregressive error accumulation via a curriculum-based CRPS finetuning strategy. Experiments show that Tyche matches or surpasses the state-of-the-art multi-step generative methods. As a current limitation, the highly efficient inference of Tyche requires a relatively high GPU memory overhead, leaving room for further research.

\section*{Acknowledgement}
This work was supported by ShenzhenLoop Area Institute (No. FP202602).

% \clearpage
\bibliography{reference.bib}
\bibliographystyle{plain}

\clearpage
\appendix
% \section*{Appendix Contents}
% \begingroup
% % \footnotesize
% \setlength{\parindent}{0pt}
% \newcommand{\apptocline}[3][0em]{%
%   \noindent\hspace*{#1}\hyperref[#2]{\ref*{#2}\hspace{0.6em}#3}%
%   \nobreak\leaders\hbox{$\mkern2mu.\mkern2mu$}\hfill\nobreak\pageref*{#2}\par}
% \newcommand{\appsec}[2]{\vspace{0.2em}\apptocline{#1}{\textbf{#2}}}
% \newcommand{\appsub}[2]{\apptocline[1.6em]{#1}{#2}}
% \newcommand{\appsubsub}[2]{\apptocline[3.2em]{#1}{#2}} % 如果需要索引 subsubsection 可使用此命令

% \appsec{app:setting}{Detailed Experimental Settings}
% \appsub{app:data}{Data Description}
% \appsub{app:baseline}{Details of Baselines and Implementations}
% % \appsubsub{app:implementation_details}{Implementation Details} % 预留了子子章节的索引，若不需要可保持注释
% \appsub{app:metrics}{Evaluation Metrics}

% \appsec{app:tyche}{Details of \ourmethod}
% \appsub{app:network}{Details of Network Architecture}
% \appsub{app:pseudocode}{Pseudocode of \ourmethod}

% \appsec{app:theory}{Theoretical Analysis and Proofs}
% \appsub{app:theory_discretization}{The Discretization Error Perspective for One-Step Transport}
% \appsub{app:theory_proof}{Proof of Theorem}
% \appsub{app:theory_bridge}{Bridging Theory and Practice: The Rationale for Stage II Design}

% \appsec{app:extended_exp}{Extended Experiments}
% \appsub{app:more_vars}{Comparison on more variables}
% \appsub{app:ablation}{Ablation study}
% \appsub{app:skill_ens}{Ensemble skills comparison}
% % \appsub{app:local_ens}{Local ensemble performance}
% \appsub{app:detailed_vis}{Detailed visualizations on more horizons and variables}

% \endgroup
% \clearpage

\section*{Appendix Contents}
\begingroup
% \footnotesize
\setlength{\parindent}{0pt}
\newcommand{\apptocline}[3][0em]{%
  \noindent\hspace*{#1}\hyperref[#2]{\ref*{#2}\hspace{0.6em}#3}%
  \nobreak\leaders\hbox{$\mkern2mu.\mkern2mu$}\hfill\nobreak\pageref*{#2}\par}
\newcommand{\appsec}[2]{\vspace{0.2em}\apptocline{#1}{\textbf{#2}}}
\newcommand{\appsub}[2]{\apptocline[1.6em]{#1}{#2}}
\newcommand{\appsubsub}[2]{\apptocline[3.2em]{#1}{#2}}

\appsec{app:setting}{Detailed Experimental Settings}
\appsub{app:data}{Data Description}
\appsub{app:baseline}{Details of Baselines and Implementations}
\appsubsub{app:deterministic_baseline}{Deterministic Baselines}
\appsubsub{app:generative_baseline}{Probabilistic and Generative Baselines}
\appsubsub{app:operational_baseline}{Operational NWP Baseline}
\appsubsub{app:implementation_details}{Implementation Details}
\appsub{app:metrics}{Evaluation Metrics}

\appsec{app:tyche}{Details of \ourmethod}
\appsub{app:network}{Details of Network Architecture}
\appsub{app:pseudocode}{Pseudocode of \ourmethod}

\appsec{app:theory}{Theoretical Analysis and Proofs}
\appsub{app:theory_discretization}{The Discretization Error Perspective for One-Step Transport}
\appsub{app:theory_proof}{Proof of Theorem}
\appsub{app:theory_bridge}{Bridging Theory and Practice: The Rationale for Stage II Design}

\appsec{app:extended_exp}{Extended Experiments}
\appsub{app:ablation}{Ablation study}
\appsub{app:more_vars}{Comparison on more variables}
\appsub{app:skill_ens}{Ensemble skills comparison}
\appsub{app:detailed_vis}{Detailed visualizations on more horizons and variables}

\endgroup
\clearpage

\section{Detailed Experimental Settings}
\label{app:setting}

\subsection{Data Description}
\label{app:data}

We use a collection of upper-air and surface-level meteorological variables as both model inputs and prediction targets. 
The full list of variables is summarized in Table~\ref{tab:data_variables}. 
For pressure-level fields, we include five atmospheric variables: Geopotential (z), Specific humidity (q), Temperature (t), U component of wind (u), and V component of wind (v).
These variables are defined on 13 standard pressure levels, namely 50, 100, 150, 200, 250, 300, 400, 500, 600, 700, 850, 925, and 1000 hPa. 
In addition, we use four single-level variables, including 10 metre u wind component (u10), 10 metre v wind component (v10), 2 metre temperature (t2m), and Mean sea level pressure (mslp).

Before training, each variable is standardized using statistics computed from the training period (1979-2017). 
The same normalization parameters are then kept fixed for all experiments to ensure consistency across different settings. 
All data are processed on a $1.5^\circ$ latitude--longitude grid. 
Following the common preprocessing strategy for this grid resolution, the southernmost latitude row is discarded from the original $1.5^\circ$ grid $(121 \times 240)$ to make the latitude dimension even.

\begin{table}[h]
\centering
\caption{Description of data variables used.}
\label{tab:data_variables}
\setlength{\tabcolsep}{10pt}  % 每一列左右两侧的内边距
\renewcommand{\arraystretch}{1.25}
\centering
\resizebox{0.8\textwidth}{!}{
\begin{tabular}{llcc}
\toprule
\textbf{Type} & \textbf{Variable name} & \textbf{Short name} & \textbf{Param ID} \\
\hline
Atmospheric & Geopotential & \texttt{z} & 129 \\
Atmospheric & Specific humidity & \texttt{q} & 133 \\
Atmospheric & Temperature & \texttt{t} & 130 \\
Atmospheric & U component of wind & \texttt{u} & 131 \\
Atmospheric & V component of wind & \texttt{v} & 132 \\
Single & 10 metre u wind component & \texttt{u10} & 165 \\
Single & 10 metre v wind component & \texttt{v10} & 166 \\
Single & 2 metre temperature & \texttt{t2m} & 167 \\
Single & Mean sea level pressure & \texttt{mslp} & 151 \\
\bottomrule
\end{tabular}
}
\end{table}

\subsection{Details of Baselines and Implementations}
\label{app:baseline}

To provide a comprehensive evaluation of \ourmethod, we compare it against a robust suite of state-of-the-art weather forecasting systems. These baselines are broadly categorized into purely deterministic ML models, modern generative/probabilistic ML models, and traditional operational numerical weather prediction (NWP).

\subsubsection{Deterministic Baselines}
\label{app:deterministic_baseline}
These models aim to minimize the mean-squared error (MSE) of the forecast but are known to oversmooth predictions at longer lead times, lacking the capability to capture the full distribution of weather uncertainties.
\begin{itemize}[leftmargin=*]
    \item \textbf{Pangu-Weather~\cite{bi2023accurate}:} A highly influential deterministic MLWP model utilizing a 3D Earth-Specific Transformer. It processes data using height-specific attention mechanisms and trains independent networks for specific lead times (\textit{e.g.}, 1h, 3h, 6h, 24h) to circumvent the error accumulation typical of high-frequency autoregressive rollouts.
    \item \textbf{GraphCast~\cite{lam2023learning}:} A state-of-the-art deterministic model based on Graph Neural Networks (GNNs). It maps the latitude-longitude grid to a multi-scale icosahedral mesh using an encode-process-decode architecture, typically generating forecasts autoregressively at 6-hour intervals. 
\end{itemize}

\subsubsection{Probabilistic and Generative Baselines}
\label{app:generative_baseline}
These models represent the forefront of global AI weather ensemble forecasting (2024--2025), aiming to produce sharp, well-calibrated ensemble members that can capture extreme events and uncertainty growth.
\begin{itemize}[leftmargin=*]
    \item \textbf{GenCast~\cite{price2025probabilistic}:} A full-space conditional autoregressive diffusion model that serves as the gold-standard ML probabilistic baseline. It learns the one-step conditional distribution and rolls out autoregressively (typically in 12-hour steps). While it achieves exceptional calibration and sharpness matching operational NWP, it relies on multiple denoising evaluations per step, making sampling computationally expensive.
    \item \textbf{ARCI~\cite{andrae2025continuous}:} A continuous lead-time diffusion model designed for temporal interpolation. Instead of costly fine-grained autoregressive rollouts, ARCI conditions the diffusion directly on lead time and correlates the driving noise (via an Ornstein-Uhlenbeck process) across times. This allows independent marginals to form a temporally consistent trajectory, combining coarse autoregressive steps (\textit{e.g.}, 24h) with cheap high-resolution (\textit{e.g.}, 6h or 1h) interpolation.
    \item \textbf{ArchesWeatherGen~\cite{couairon2024archesweather}:} A computationally efficient, two-stage residual generative framework. It first utilizes a deterministic backbone (ArchesWeather, featuring Cross-Level Attention) to predict the mean state. Subsequently, a flow-matching generative model is trained to predict the normalized residuals between the deterministic output and the ERA5 ground truth. This explicitly decouples mean prediction from distributional correction.
    \item \textbf{Swift~\cite{stock2025swift}:} A highly efficient 1-NFE consistency model. Swift trains a probability-flow/diffusion-style model into a single-step consistency model using trigflow. A key innovation is its autoregressive multi-step fine-tuning utilizing the Continuous Ranked Probability Score. It represents the state-of-the-art for ultra-fast, single-step ensemble generation.
    \item \textbf{NeuralGCM-ENS~\cite{kochkov2024neural}:} A hybrid approach bridging traditional physics and machine learning. It relies on a differentiable general circulation model (dynamical core) combined with neural networks that parameterize unresolved sub-grid physics. The ensemble variant generates multiple trajectories via initial perturbations, providing a strong, physically-constrained probabilistic baseline.
\end{itemize}

\subsubsection{Operational NWP Baseline}
\label{app:operational_baseline}
\begin{itemize}[leftmargin=*]
    \item \textbf{IFS-ENS~\cite{rasp2024weatherbench}:} The ECMWF Integrated Forecasting System Ensemble. This is the gold-standard operational numerical weather prediction baseline. It relies on solving physical equations explicitly and uses Singular Vectors and Ensemble of Data Assimilations (EDA) to perturb initial conditions, typically generating a 50-member ensemble with plus one unperturbed control.
\end{itemize}

\subsubsection{Implementation Details}
\label{app:implementation_details}

In our experiments, \ourmethod is trained on the ERA5 dataset at a spatial resolution of $1.5^\circ$, totally utilizing 69 atmospheric and surface variables. The model is trained on 8 $\times$ 80GB NVIDIA H100 GPUs. During inference, an ensemble of 20 members is generated parallelly on a single GPU. 
% Compared to multi-step diffusion baselines like GenCast, \ourmethod requires only 1 NFE per forecast step, taking approximately 8 seconds to generate a 15-day (60 iteration) trajectory, thereby offering a highly favorable compute-skill Pareto frontier.

\paragraph{Architecture of \ourmethod.} 
The backbone of \ourmethod is a non-hierarchical Swin Transformer configured with 12 transformer blocks. Each block features a hidden dimension of 1024 and 12 attention heads. Spatial interactions are restricted to local windows of size $10 \times 10$, which are cyclically shifted by $5 \times 5$ in alternating blocks. To effectively capture spatial dependencies and temporal dynamics, we employ learned absolute positional embeddings combined with 2D axial Rotary Positional Embeddings (RoPE) within the windows. Furthermore, the architecture utilizes pre-RMSNorm paired with adaptive modulation (generating scale, shift, and gating factors conditioned on the time steps) to stabilize the generative learning process. In this configuration, the model contains approximately 230 MB trainable parameters.

\paragraph{Training Pipeline of \ourmethod.} 
The training pipeline is meticulously divided into two stages to decouple deterministic representation learning from probabilistic calibration:

\ding{224} \textbf{Stage I (One-step flow pretraining):} The model is trained from scratch to learn the underlying vector field using an MSE-based objective with 300 epochs. We use the AdamW optimizer with a weight decay of $1e^{-4}$ and an initial learning rate of $1e^{-4}$, which is scheduled by a cosine annealing policy decaying to a minimum learning rate of $1e^{-6}$.

\ding{224} \textbf{Stage II (Curriculum probabilistic calibration):} To explicitly calibrate the uncertainty and ensure long-rollout stability, we perform curriculum autoregressive training using an empirical CRPS loss. The training utilizes a small ensemble size of 2 to maintain computational tractability. The autoregressive rollout horizon progressively increases from 1 to 2 steps, with each curriculum stage trained for 15 epochs. The learning rate is reduced and fixed at $1e^{-5}$ for this stage.

\paragraph{Baseline Configurations and Fairness.} 
To ensure a rigorous and fair evaluation, the implementations of our selected baselines are handled as follows:

\ding{224} \textbf{Trained from Scratch:}  For Pangu, GraphCast, ARCI, Swift, and ArchesWeatherGen, we utilize their official codebases and train them from scratch on the exact same $1.5^\circ$ ERA5 dataset split as \ourmethod. All training for these baselines is uniformly conducted on eight 80GB NVIDIA H100 GPUs.

\ding{224} \textbf{Downloaded Operational/Large-Scale References:} For GenCast, NeuralGCM-ENS, and IFS-ENS, training from scratch is operationally restricted. We directly download their official saved inference data for the corresponding evaluation years and variables from WeatherBench2~\cite{rasp2024weatherbench}.

Across all probabilistic methods and operational references, the evaluated ensemble number is strictly aligned to 20 members to ensure fair comparison of generative diversity and calibration metrics.

\subsection{Evaluation Metrics}
\label{app:metrics}

To comprehensively evaluate the performance of our probabilistic forecasts, we employ standard metrics from meteorology and uncertainty quantification, including the Latitude-weighted Root Mean Square Error (RMSE), Ensemble Spread, Spread-to-Skill Ratio (SSR), and Continuous Ranked Probability Score (CRPS). 

To ensure fair evaluation across the globe, all spatial aggregations are weighted by latitude. Let $M$ be the total number of grid cells. For a grid cell $i$ at latitude $\theta_i$, the normalized area-based latitude weight $w_i$ is defined as:
\begin{equation}
    w_i = \frac{\cos(\theta_i)}{\frac{1}{M}\sum_{j=1}^M \cos(\theta_j)},
\end{equation}
such that the weights average to 1 over the entire grid. 

In the following formulations, let $N$ denote the total number of forecast initialization times (indexed by $n$), and $K$ denote the ensemble size (indexed by $k$). Let $y_{i,n}$ represent the ground truth for grid cell $i$ at time $n$, $x_{i,n}^k$ represent the corresponding prediction of the $k$-th ensemble member, and $\bar{x}_{i,n} = \frac{1}{K}\sum_{k=1}^K x_{i,n}^k$ denote the ensemble mean.

\paragraph{Latitude-weighted RMSE (Skill).} 
RMSE measures the deterministic accuracy of the forecast. For ensemble models, it is calculated based on the ensemble mean:
\begin{equation}
    \text{RMSE} = \frac{1}{N} \sum_{n=1}^N \sqrt{ \frac{1}{M} \sum_{i=1}^M w_i \left( y_{i,n} - \bar{x}_{i,n} \right)^2 }.
\end{equation}

\paragraph{Ensemble Spread.}
Spread quantifies the variability or uncertainty captured within the ensemble members. It is calculated as the root mean square of the ensemble variance:
\begin{equation}
    \text{Spread} = \frac{1}{N} \sum_{n=1}^N \sqrt{ \frac{1}{M} \sum_{i=1}^M w_i \left( \frac{1}{K-1} \sum_{k=1}^K (x_{i,n}^k - \bar{x}_{i,n})^2 \right) }.
\end{equation}

\paragraph{Spread-to-Skill Ratio (SSR).}
Ideally, a well-calibrated probabilistic forecast should achieve a balance where the ensemble spread is proportional to the RMSE error of the ensemble mean. The SSR evaluates this reliability, with values close to 1 indicating optimal effective uncertainty estimation:
\begin{equation}
    \text{SSR} = \sqrt{\frac{K+1}{K}} \frac{\text{Spread}}{\text{RMSE}},
\end{equation}
where $\sqrt{(K+1)/K}$ is a statistical correction factor for finite ensemble sizes.

\paragraph{Continuous Ranked Probability Score (CRPS).}
CRPS is a widely used metric for probabilistic forecasting that compares the cumulative distribution function (CDF) of the predicted ensemble against the observation. It penalizes errors in both the location and the shape of the predicted distribution. Following the standard in WeatherBench2, we utilize the unbiased estimator of CRPS:
\begin{equation}
    \text{CRPS} = \frac{1}{N} \sum_{n=1}^N \frac{1}{M} \sum_{i=1}^M w_i \left( \frac{1}{K} \sum_{k=1}^K |x_{i,n}^k - y_{i,n}| - \frac{1}{2K(K-1)} \sum_{k=1}^K \sum_{k'=1}^K |x_{i,n}^k - x_{i,n}^{k'}| \right).
\end{equation}

\section{Details of Tyche}
\label{app:tyche}

\subsection{Details of Network Architecture}
\label{app:network}

\begin{figure*}[h]
\centering
\includegraphics[width=1.0\linewidth]{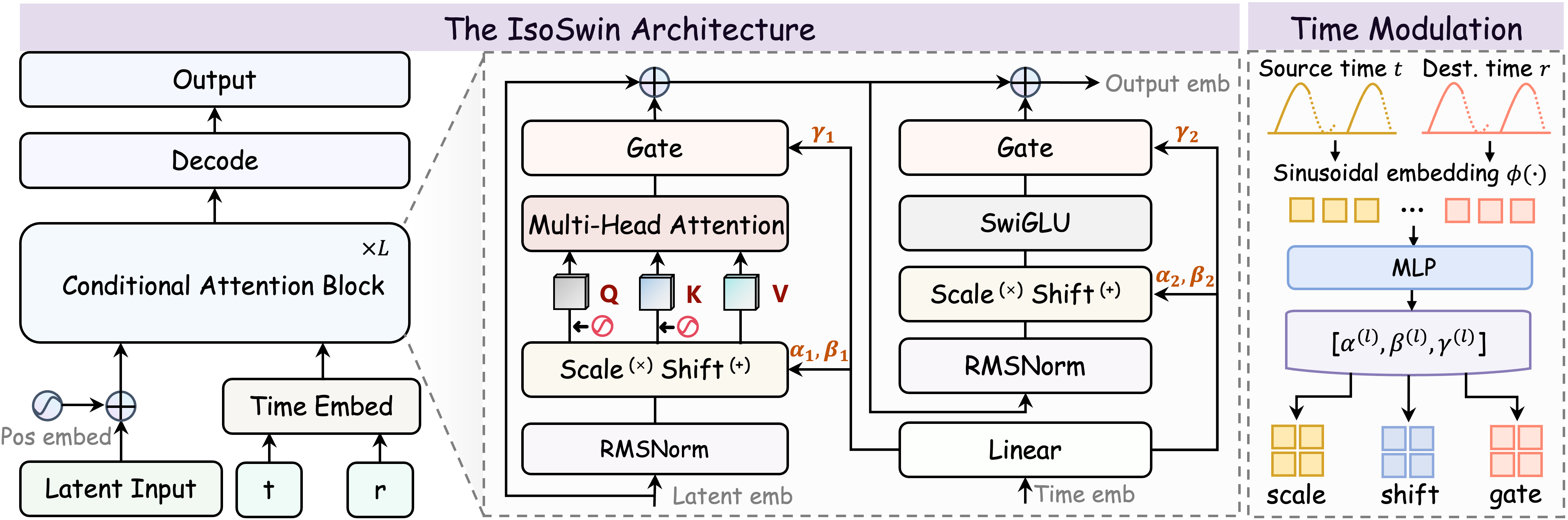}
\vspace{-10pt}
\caption{The details of network architecture. }
\label{fig:model_appen}
\vspace{-5pt}
\end{figure*}

The transport field in Eq.~\eqref{eq:lflow} is parameterized by an isotropic Swin-style transformer that operates on the concatenation of the conditioning field and the noisy target state. In formulation:
\begin{equation}
z_{\mathrm{in}} = [c; z_t] \in \mathbb{R}^{2C \times H \times W},
\end{equation}
partition $z_{\mathrm{in}}$ into non-overlapping patches of size $P \times P$, and project them into tokens:
\begin{equation}
z^{(0)}
=
\operatorname{Linear}\!\big(\operatorname{Patchify}(z_{\mathrm{in}})\big)
+
P_{\mathrm{abs}},
\label{eq:patch}
\end{equation}
where $P_{\mathrm{abs}}$ is a learned absolute positional embedding that provides fixed geographic context.

We encode the source time $t$ and destination time $r$ by sinusoidal embeddings followed by an MLP:
\begin{equation}
t_{\mathrm{emb}} = f_t(\phi(t)) +  f_r(\phi(r)).
\label{eq:temb}
\end{equation}
This temporal embedding modulates each transformer block through adaptive normalization. 
For the $l$-th block, we predict scale, shift, and residual gates from $t_{\mathrm{emb}}$:
\begin{equation}
[\alpha^{(l)}, \beta^{(l)}, \gamma^{(l)}]
=
W_{\mathrm{mod}}^{(l)} t_{\mathrm{emb}} + b_{\mathrm{mod}}^{(l)}.
\label{eq:mod}
\end{equation}
The modulated features are:
\begin{equation}
\mathcal{M}^{(l)}(z)
=
\operatorname{RMSNorm}(z)
\odot
(1+\alpha^{(l)})
+
\beta^{(l)}.
\label{eq:adarms}
\end{equation}

To preserve fine-scale meteorological structure, we avoid hierarchical downsampling and keep an isotropic resolution throughout the backbone. 
Self-attention is restricted to local windows, while consecutive blocks alternate between regular and shifted windows to exchange information across window boundaries. 
In addition, we apply 2D axial rotary positional embeddings (RoPE) to the query and key tokens inside each window to provide relative spatial awareness for advection and vortex-like motion. 
The $l$-th block is written as:
\begin{align}
\tilde{z}^{(l)}
&=
\hat{z}^{(l)}
+
\boldsymbol{\gamma}^{(l)}_{\mathrm{att}}
\odot
\operatorname{WAttn}_{\mathrm{RoPE}}
\big(
\mathcal{M}^{(l)}_{\mathrm{att}}(\hat{z}^{(l)})
\big),
\\
z^{(l)}
&=
\mathcal{S}^{-1}
\Big(
\tilde{z}^{(l)}
+
\boldsymbol{\gamma}^{(l)}_{\mathrm{ffn}}
\odot
\operatorname{SwiGLU}
\big(
\mathcal{M}^{(l)}_{\mathrm{ffn}}(\tilde{z}^{(l)})
\big)
\Big),
\label{eq:isoblock}
\end{align}
where $\mathcal{S}$ denotes the cyclic shift operator. 
The residual gates are zero-initialized to stabilize optimization in the early stage of training.

After the final block, a lightweight decoder maps the tokens back to the physical grid and outputs the velocity field:
\begin{equation}
u_{\theta}(z_s, s, c, r)
=
\operatorname{Unpatchify}\!\big(W_{\mathrm{out}}z^{(L)}\big).
\label{eq:decoder}
\end{equation}
We additionally attach a variance head that predicts a non-negative scale field $\sigma_{\theta}$, which is used for uncertainty modulation during finetuning and inference.

\subsection{Pseudocode of \ourmethod}
\label{app:pseudocode}

The detailed pseudocode of \ourmethod is illustrated in Algorithm~\ref{alg:tyche_workflow}.

\begin{algorithm}[htbp]
\caption{Training and Inference Workflow of \ourmethod{}}
\label{alg:tyche_workflow}
\begin{algorithmic}[1]
\Require Training dataset $\mathcal{D} = \{(x^\tau, x^{\tau+1}, \dots)\}$, model parameters $\theta$, ensemble size $K$.

\Statex \textbf{/* Stage I: One-step Transport Pretraining */}
\For{each pretraining iteration}
    \State Sample $(x^\tau, x^{\tau+1}) \sim \mathcal{D}$, noise $\epsilon \sim \mathcal{N}(0, I)$, and times $r < t \in [0, 1]$.
    \State Construct intermediate state $z_t \gets (1-t)x^{\tau+1} + t\epsilon$ and target $v_t \gets \epsilon - x^{\tau+1}$.
    \State Compute total derivative $\frac{d}{dt}u_\theta^\star$ via Jacobian-Vector Product (JVP).
    \State Calculate rectified target $u_{\mathrm{tgt}} \gets v_t - (t-r) \frac{d}{dt}u_\theta^\star$. \hfill $\triangleright$ Eq.~\ref{eq:utgt}
    \State Update $\theta$ by minimizing $\mathcal{L}_{\mathrm{ave}} = \text{MSE}\big(u_\theta(z_t, r, t, x^\tau), \operatorname{sg}(u_{\mathrm{tgt}})\big)$. \hfill $\triangleright$ Eq.~\ref{eq:lflow}
\EndFor

\Statex \textbf{/* Stage II: Curriculum Probabilistic Calibration */}
\For{each curriculum stage with rollout horizon $R_m$}
    \For{each finetuning iteration}
        \State Sample initial condition $x^\tau$ and target state $x^{\tau+R_m}$ from $\mathcal{D}$.
        \State Initialize $K$ parallel paths: $c_k^0 \gets x^\tau$ for $k \in \{1,\dots,K\}$.
        
        \For{lead time $l = 1$ to $R_m$} \Comment{Autoregressive 1-NFE Rollout}
            \State Sample $\epsilon_k^l \sim \mathcal{N}(0, I)$ and generate $\hat{x}_k^{\tau+l} \gets \epsilon_k^l - u_\theta(\epsilon_k^l, 0, 1, c_k^{l-1})$.
            \State Update conditioning: $c_k^l \gets \hat{x}_k^{\tau+l}$.
        \EndFor
        
        \State Update $\theta$ by minimizing empirical $\mathcal{L}_{\mathrm{CRPS}}$ on terminal states $\{\hat{x}_k^{\tau+R_m}\}$. \hfill $\triangleright$ Eq.~\ref{eq:empirical_crps}
    \EndFor
\EndFor

\Statex \textbf{/* Inference: 1-NFE Ensemble Generation */}
\State Sample $K$ independent Gaussian noises $\{\epsilon_1, \dots, \epsilon_K\} \sim \mathcal{N}(0, I)$.
\For{member $k = 1$ to $K$ \textbf{in parallel}}
    \State Generate forecast sample: $\hat{x}_k^{\tau+1} \gets \epsilon_k - u_\theta(\epsilon_k, 0, 1, x^\tau)$. \hfill $\triangleright$ Eq.~\ref{eq:ensemble_inference}
\EndFor
\State \Return Generated forecast ensemble $\hat{X} = \{\hat{x}_1^{\tau+1}, \dots, \hat{x}_K^{\tau+1}\}$.
\end{algorithmic}
\end{algorithm}

\section{Theoretical Analysis and Proofs}
\label{app:theory}

We analyze how errors in the learned conditional transition distribution propagate during autoregressive forecasting. We first distinguish approximation error in average-velocity transport from numerical integration error. We then derive a rollout error bound under a Lipschitz assumption on the true transition kernel and relate the local kernel discrepancy to a pathwise rectification residual. Finally, we discuss the implications and limitations of this analysis for Stage II finetuning.

\subsection{Approximation Error in One-Step Transport}
\label{app:theory_discretization}

Standard flow matching parameterizes an instantaneous velocity field, whose associated ordinary differential equation is typically solved by numerical integration. In contrast, \ourmethod{} parameterizes an interval-averaged velocity. For the exact average-velocity field $u^\star$, the endpoints of the corresponding transport trajectory satisfy
\begin{equation}
    z_r = z_s-(s-r)u^\star(z_s,r,s,c), \qquad 0 \leq r < s \leq 1.
    \label{eq:app_exact_transport}
\end{equation}
Using a learned field $u_\theta$ gives $\widehat z_r=z_s-(s-r)u_\theta(z_s,r,s,c)$, and hence
\begin{equation}
    \widehat z_r-z_r = -(s-r) \left( u_\theta(z_s,r,s,c) - u^\star(z_s,r,s,c) \right).
    \label{eq:app_transport_approximation}
\end{equation}
Thus, the endpoint error arises from approximation of the average-velocity field. The update should not be identified with a single Euler step applied to an instantaneous velocity field, nor does it incur an unavoidable Euler truncation error when the average velocity is exact.

During autoregressive forecasting, subsequent transitions are conditioned on generated states. Their distribution may differ from that of the observed conditioning states used in Stage I. The following analysis describes how local transition discrepancies and sensitivity to conditioning states contribute to the resulting rollout error.

\subsection{Rollout Error Bound and Proof}
\label{app:theory_proof}

We consider the Markov formulation used in the main text. Let $\mathcal X=\mathbb R^M$ denote the vectorized forecast state space. To reflect spatial weighting, equip this space with the weighted $L_1$ norm
\begin{equation}
    \|x\|_{1,w} \coloneqq \sum_{i=1}^{M} w_i |x_i|, \qquad w_i>0,
    \label{eq:app_weighted_norm}
\end{equation}
where the weights are fixed. Let $W_{1,w}$ be the Wasserstein-1 distance induced by this ground metric.

For a fixed conditioning state $c$, define the one-step sampler and its transition kernel by
\begin{equation}
    \begin{aligned}
        g_\theta(\epsilon,c) &\coloneqq \epsilon-u_\theta(\epsilon,0,1,c), \qquad \epsilon\sim\mathcal N(0,I), \\
        K_\theta(\cdot\mid c) &\coloneqq \operatorname{Law}\bigl(g_\theta(\epsilon,c)\bigr).
    \end{aligned}
    \label{eq:app_model_kernel}
\end{equation}
Let $K^\star(\cdot\mid c)$ denote the true transition kernel. The local kernel discrepancy is
\begin{equation}
    d_\theta(c) \coloneqq W_{1,w} \left( K_\theta(\cdot\mid c), K^\star(\cdot\mid c) \right).
    \label{eq:app_exact_kernel_gap}
\end{equation}

We also define a residual-based upper bound. For each fixed $c$, independently sample $x\sim K^\star(\cdot\mid c)$ and $\epsilon\sim\mathcal N(0,I)$, and set
\begin{equation}
    z_s^{x,\epsilon} = (1-s)x+s\epsilon, \qquad v^{x,\epsilon} = \epsilon-x.
    \label{eq:app_linear_path}
\end{equation}
The sample-path rectification residual at destination time $r=0$ is
\begin{equation}
    \rho_\theta^{x,\epsilon}(s,c) \coloneqq\; u_\theta(z_s^{x,\epsilon},0,s,c) - v^{x,\epsilon} + s\left[ \partial_s u_\theta(z_s^{x,\epsilon},0,s,c) + J_z u_\theta(z_s^{x,\epsilon},0,s,c) v^{x,\epsilon} \right].
    \label{eq:app_sample_residual}
\end{equation}
Here, $\partial_s$ differentiates the explicit source-time argument while holding the spatial argument fixed. Define
\begin{equation}
    \delta_\theta(c) \coloneqq \int_0^1 \mathbb E_{x,\epsilon\mid c} \left[ \|\rho_\theta^{x,\epsilon}(s,c)\|_{1,w} \right] \,\mathrm ds.
    \label{eq:app_residual_upper_bound}
\end{equation}

We assume that the kernels are measurable, that the transition and rollout distributions considered below have finite first moments, and that all residual expectations are finite. For each fixed $c$, we assume sufficient regularity of $u_\theta$ for the chain rule and the fundamental theorem of calculus along the paths below. In particular, the field is continuous at the diagonal $s=r$, continuously differentiable along the relevant paths, and its total derivative is integrable. Continuity at the diagonal ensures
\begin{equation}
    \lim_{s\downarrow r} (s-r)u_\theta(z_s,r,s,c)=0.
    \label{eq:app_diagonal_boundary}
\end{equation}

\textbf{Theorem 1 (\textit{State-sensitive rollout error bound}).} \textit{Assume that the true transition kernel satisfies
\begin{equation}
    W_{1,w} \left( K^\star(\cdot\mid c), K^\star(\cdot\mid c') \right) \leq \Lambda\|c-c'\|_{1,w}
    \label{eq:app_kernel_lipschitz}
\end{equation}
for all conditioning states in the common state space, including those visited by the model, where $\Lambda\geq0$. Let $\mu_h^\theta$ and $\mu_h^\star$ be the model and true rollout distributions, respectively, with
\[
    \mu_{h+1}^\theta=\mu_h^\theta K_\theta, \qquad \mu_{h+1}^\star=\mu_h^\star K^\star, \qquad \mu_0^\theta=\mu_0^\star.
\]
Under the regularity and integrability assumptions above, for every $H\geq1$, }
\begin{equation}
    W_{1,w} \left( \mu_H^\theta,\mu_H^\star \right) \leq \sum_{j=0}^{H-1} \Lambda^{H-1-j} \mathbb E_{c\sim\mu_j^\theta} \left[d_\theta(c)\right] \leq \sum_{j=0}^{H-1} \Lambda^{H-1-j} \mathbb E_{c\sim\mu_j^\theta} \left[\delta_\theta(c)\right].
    \label{eq:app_theorem1_final}
\end{equation}
\textit{The coefficient for $j=H-1$ is understood to be $1$, including when $\Lambda=0$. }

\begin{proof}
The proof first establishes a pathwise endpoint identity, then bounds the local kernel discrepancy, and finally propagates this discrepancy through the autoregressive chain.

\textbf{Step 1: Residual-to-endpoint identity.} Fix a conditioning state $c$, a destination time $r$, and a continuously differentiable path $z_s$ with velocity $v_s=\mathrm dz_s/\mathrm ds$. For a candidate average-velocity field $u$, define its total derivative along this path by
\begin{equation}
    D_su(z_s,r,s,c) \coloneqq \partial_su(z_s,r,s,c) + J_zu(z_s,r,s,c)v_s.
\end{equation}
The associated pathwise residual is
\begin{equation}
    \rho_u(s) \coloneqq u(z_s,r,s,c)-v_s +(s-r)D_su(z_s,r,s,c).
    \label{eq:app_residual}
\end{equation}
Let $\widehat z_r(s)=z_s-(s-r)u(z_s,r,s,c)$. Differentiation gives
\begin{equation}
    \frac{\mathrm d}{\mathrm ds}\widehat z_r(s) = -\rho_u(s).
\end{equation}
The diagonal boundary condition implies $\widehat z_r(r)=z_r$. Integrating from $r$ to $s$ therefore yields
\begin{equation}
    \widehat z_r(s)-z_r = -\int_r^s\rho_u(a)\,\mathrm da.
    \label{eq:app_endpoint_identity}
\end{equation}
By the triangle inequality,
\begin{equation}
    \|\widehat z_r(s)-z_r\|_{1,w} \leq \int_r^s\|\rho_u(a)\|_{1,w}\,\mathrm da.
    \label{eq:app_endpoint_bound}
\end{equation}

\textbf{Step 2: From endpoint error to kernel discrepancy.} Apply Step 1 to the linear path in Eq.~\eqref{eq:app_linear_path}, with $u=u_\theta$, $r=0$, and $s=1$. Its endpoints are $z_0=x$ and $z_1=\epsilon$, while the predicted destination is $\widehat x=g_\theta(\epsilon,c)$. Consequently,
\begin{equation}
    \|\widehat x-x\|_{1,w} \leq \int_0^1 \|\rho_\theta^{x,\epsilon}(s,c)\|_{1,w} \,\mathrm ds.
    \label{eq:app_endpoint_l1}
\end{equation}
The pair $(\widehat x,x)$ is a valid coupling of $K_\theta(\cdot\mid c)$ and $K^\star(\cdot\mid c)$. Since the Wasserstein distance is the infimum of the expected ground cost over all couplings,
\begin{equation}
    d_\theta(c) \leq \mathbb E_{x,\epsilon\mid c} \left[ \|\widehat x-x\|_{1,w} \right] \leq \int_0^1 \mathbb E_{x,\epsilon\mid c} \left[ \|\rho_\theta^{x,\epsilon}(s,c)\|_{1,w} \right] \,\mathrm ds =\delta_\theta(c).
    \label{eq:app_local_kernel_bound}
\end{equation}

\textbf{Step 3: Autoregressive error propagation.} Write $e_h=W_{1,w}(\mu_h^\theta,\mu_h^\star)$. The triangle inequality gives:
\begin{equation}
    e_{h+1} \leq W_{1,w} \left( \mu_h^\theta K_\theta, \mu_h^\theta K^\star \right) + W_{1,w} \left( \mu_h^\theta K^\star, \mu_h^\star K^\star \right).
    \label{eq:app_triangle}
\end{equation}
The mixture inequality for Wasserstein distance bounds the first term by
\begin{equation}
    W_{1,w} \left( \mu_h^\theta K_\theta, \mu_h^\theta K^\star \right) \leq \mathbb E_{c\sim\mu_h^\theta} \left[d_\theta(c)\right].
    \label{eq:app_model_gap}
\end{equation}
For the second term, let $\pi_h$ be an optimal coupling of $\mu_h^\theta$ and $\mu_h^\star$. Using Eq.~\eqref{eq:app_kernel_lipschitz},
\begin{equation}
    \begin{aligned}
        W_{1,w} \left( \mu_h^\theta K^\star, \mu_h^\star K^\star \right) &\leq \int W_{1,w} \left( K^\star(\cdot\mid c), K^\star(\cdot\mid c') \right) \,\pi_h(\mathrm dc,\mathrm dc') \\
        &\quad\leq \Lambda \int \|c-c'\|_{1,w} \,\pi_h(\mathrm dc,\mathrm dc') = \Lambda e_h.
    \end{aligned}
    \label{eq:app_true_gap}
\end{equation}
Combining these inequalities yields
\begin{equation}
    e_{h+1} \leq \Lambda e_h + \mathbb E_{c\sim\mu_h^\theta} \left[d_\theta(c)\right].
    \label{eq:app_recursion}
\end{equation}
Since $e_0=0$, unrolling the recursion proves the first inequality in Eq.~\eqref{eq:app_theorem1_final}. Applying $d_\theta(c)\leq\delta_\theta(c)$ proves the second.
\end{proof}

\textbf{Interpretation of the residual upper bound.} The quantity $d_\theta(c)$ is the actual discrepancy between the conditional transition distributions, whereas $\delta_\theta(c)$ is a sufficient, potentially loose upper bound. In Step 2, $x$ and $\widehat x$ are independent conditional on $c$, because the verifying target and sampling noise are drawn independently. This coupling is valid but generally not optimal. For a non-degenerate true distribution, its expected cost can remain positive even when $K_\theta(\cdot\mid c)=K^\star(\cdot\mid c)$. Accordingly, $\delta_\theta(c)$ need not vanish for a distributionally correct sampler and should not be interpreted as an exact measure of distributional error. The residual certificate also uses the fixed destination $r=0$. A small Stage I loss averaged over sampled $(r,s)$ does not, by itself, establish a small residual on this boundary. Such a conclusion would require additional control of the time dependence or explicit supervision of the relevant boundary slice.

\textbf{Interpretation of the sensitivity factor.} The constant $\Lambda$ bounds the sensitivity of the true transition kernel in the chosen Wasserstein metric. It is not identified with a Lyapunov exponent. When $\Lambda>1$, the bound permits amplification of earlier transition discrepancies; it does not imply that the actual forecast error necessarily grows at that rate. For example, if $\mathbb E_{c\sim\mu_j^\theta}[d_\theta(c)]\leq\bar d$ for all steps under consideration, then:
\begin{equation}
    e_H \leq \bar d \sum_{k=0}^{H-1}\Lambda^k.
    \label{eq:app_uniform_local_error}
\end{equation}
This gives a horizon-independent bound $e_H\leq\bar d/(1-\Lambda)$ when $\Lambda<1$, a bound $e_H\leq H\bar d$ when $\Lambda=1$, and a potentially exponentially growing upper bound when $\Lambda>1$. The theorem therefore does not establish long-horizon stability without additional assumptions.

\subsection{Bridging Theory and Practice: The Rationale for Stage II Design}
\label{app:theory_bridge}

Theorem 1 evaluates local kernel discrepancies under the model-generated state distributions $\mu_j^\theta$. This highlights the importance of prediction quality under generated conditioning states, beyond the observed states used during Stage I. Motivated by this observation, Stage II uses autoregressive ensemble rollouts and probabilistic supervision at their terminal forecast states.

\textbf{Marginal Calibration with Fair CRPS.} For a fixed initial condition, forecast lead time, and scalar forecast component, let $F_\theta$ and $G$ denote the predictive and true cumulative distribution functions. Given $K\geq2$ conditionally independent ensemble samples $X^{1:K}\sim F_\theta$ and an observation $Y\sim G$, independent of the sampling noise, the fair CRPS estimator is
\begin{equation}
    \widehat{\mathrm{CRPS}}_{\mathrm{fair}}(X^{1:K},Y) = \frac{1}{K}\sum_{k=1}^{K}|X^k-Y| - \frac{1}{2K(K-1)}\sum_{\substack{k,k'=1\\k\neq k'}}^{K}|X^k-X^{k'}|.
    \label{eq:fair_crps_bridge}
\end{equation}
This is an unbiased estimator of the population CRPS $\mathrm{CRPS}(F_\theta,Y)$, conditional on $Y$ and the initial condition. For distributions with finite first moments, the expected excess score satisfies
\begin{equation}
    \mathbb E_{X^{1:K},Y}\left[\widehat{\mathrm{CRPS}}_{\mathrm{fair}}(X^{1:K},Y)\right] - \mathbb E_{Y\sim G}\left[\mathrm{CRPS}(G,Y)\right] = \int_{\mathbb R}\left(F_\theta(z)-G(z)\right)^2\,\mathrm dz.
    \label{eq:crps_excess_bridge}
\end{equation}
This follows from the CDF representation
\[
    \mathrm{CRPS}(F,y) = \int_{\mathbb R}\left(F(z)-\mathbf 1\{y\leq z\}\right)^2\,\mathrm dz.
\]
Thus, the expected score is minimized when the predictive marginal matches the true marginal. Applying this score to individual forecast components and aggregating with latitude weights provides a tractable objective for marginal distributional accuracy.

The resulting objective is not equivalent to the joint Wasserstein distance in Theorem 1. Matching all scored marginals does not, by itself, identify the spatial or cross-variable dependence structure. Furthermore, terminal CRPS evaluates the forecast distribution produced by a rollout; it does not directly estimate $d_\theta(c)$ or the residual upper bound $\delta_\theta(c)$ at intermediate generated states. Its role is to provide probabilistic supervision for rollout outcomes, consistent with the motivation to train beyond observed conditioning states.

\textbf{Short Rollouts for Training on Generated States.} We use a curriculum with rollout horizons $R=1$ and $R=2$ to introduce autoregressive feedback while keeping training computation and memory requirements manageable. The one-step stage supervises forecasts conditioned on observed states. In the two-step stage, the second transition is conditioned on the model's own first-step prediction, and the terminal CRPS evaluates the resulting two-step forecast distribution. Thus, the second transition encounters conditioning states distributed according to $\mu_1^\theta$, while $\mu_2^\theta$ describes the terminal predictions.

The amplification factors in Theorem 1 motivate attention to errors along generated trajectories, but they do not determine the optimal training horizon or characterize gradient variance. The short-rollout curriculum is therefore a practical training choice. Its effectiveness must be assessed empirically, and the theorem alone does not guarantee that short-rollout finetuning controls errors at arbitrary forecast horizons.

\section{Extended Experiments}
\label{app:extended_exp}

\subsection{Ecternal ablation study}
\label{app:ablation}

\begin{figure*}[h]
\centering
\includegraphics[width=1.0\linewidth]{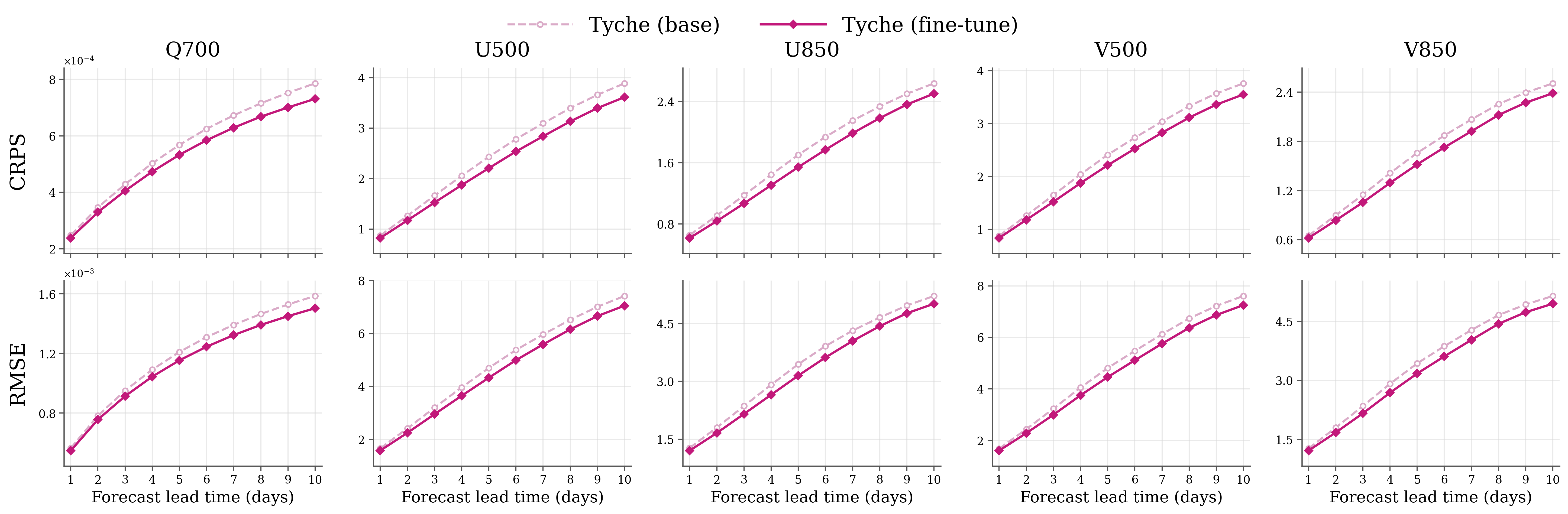}
\vspace{-15pt}
\caption{Ablation study on the ensemble finetuning. We compare the latitude-weighted CRPS (top) and RMSE (bottom) of the pretrained \ourmethod \textit{(base)} versus the fully trained \ourmethod \textit{(fine-tune)} across five upper-air variables over a 10-day horizon.
}
\label{fig:ap_abla}
\vspace{-10pt}
\end{figure*}

To validate the necessity of our proposed two-stage training strategy, we conduct an ablation study isolating the impact of the Stage II curriculum probabilistic calibration. Figure~\ref{fig:ap_abla} compares the predictive performance of the pretrained base model (\ourmethod \textit{(base)}) against the fully finetuned model (\ourmethod \textit{(fine-tune)}) over a 10-day autoregressive rollout. We report both CRPS and RMSE across representative upper-air kinematic and thermodynamic variables (Q700, U500, U850, V500, V850).

As illustrated in Figure~\ref{fig:ap_abla}, the base model is significantly outperformed by its fine-tuned counterpart, suffering from increasingly severe compounding errors during inference. Conversely, the fine-tuned model effectively constrains this error accumulation. Such a performance gap provides strong empirical corroboration for our theoretical analysis. Specifically, the CRPS-based curriculum fine-tuning functions as an \textit{amortized correction mechanism}; it exposes the model to its own autoregressive drift (on-policy), thereby enforcing the temporal self-consistency required for stable long-term probabilistic forecasting.

\subsection{Comparison on more variables}
\label{app:more_vars}

To comprehensively evaluate predictive skill across diverse meteorological fields, we report the latitude-weighted RMSE for 4 surface and 10 upper-air variables. We evaluate under two distinct settings: single-trajectory deterministic skill (Table \ref{tab:extended_rmse}) and ensemble forecasting capability (Table \ref{tab:swift_vs_tyche_rmse}).

\textbf{Single-Trajectory Predictive Skill.} Table \ref{tab:extended_rmse} compares the deterministic model GraphCast against a \textit{single ensemble member} from the probabilistic models (ArchesGen and \ourmethod). As expected, GraphCast holds a slight advantage in short-term (1-day) forecasts, particularly for wind components. However, a single member of \ourmethod remains highly competitive early on and excels in thermodynamic variables (e.g., t2m, q7). Notably, at medium-range horizons (3 to 10 days), \ourmethod's single trajectory systematically outperforms both GraphCast and ArchesWeatherGen across the vast majority of variables. This demonstrates that \ourmethod produces highly realistic and stable individual atmospheric trajectories without suffering from severe autoregressive error accumulation.

\textbf{Ensemble Forecasting Capability.} Table \ref{tab:swift_vs_tyche_rmse} specifically assesses uncertainty modeling by comparing the RMSE of the \textit{20-member ensemble mean} for probabilistic models. Under this rigorous ensemble evaluation, \ourmethod exhibits absolute superiority, achieving the lowest RMSE across \textit{all} 14 variables at \textit{every} evaluated lead time (1 to 10 days). This consistent across-the-board improvement confirms that \ourmethod's one-step generative transport not only maintains trajectory diversity but also guarantees a highly accurate and calibrated central predictive tendency for complex global weather fields.

\begin{table}[ht]
\centering
% \vspace{-10pt}
\caption{Extended evaluation of latitude-weighted RMSE for surface and upper-air variables across different forecast lead times. Best results are highlighted in \textbf{bold}.}
\label{tab:extended_rmse}
\setlength{\tabcolsep}{4.5pt} 
\resizebox{1.0\textwidth}{!}{
\begin{tabular}{@{}ll *{4}{c} *{10}{c}@{}}
\toprule
\multirow{2}{*}{\textbf{Lead Time}} & \multirow{2}{*}{\textbf{Model}} & \multicolumn{4}{c}{\textbf{Surface Variables}} & \multicolumn{10}{c}{\textbf{Upper-Air Variables (Pressure Levels)}} \\
\cmidrule(lr){3-6} \cmidrule(l){7-16}
& & u10 & v10 & t2m & mslp & z7 & z9 & q7 & q9 & t8 & t10 & U7 & U9 & V7 & V9 \\
\midrule
\multirow{3}{*}{1-Day} 
& GraphCast      & \textbf{0.96} & \textbf{1.01} & 1.24 & 1.09 & \textbf{0.80} & 0.71 & 3.23 & 6.39 & 0.69 & 0.98 & \textbf{1.81} & \textbf{1.57} & \textbf{1.92} & \textbf{1.56} \\
& ArchesGen          & 1.11 & 1.16 & 1.25 & 1.27 & 1.22 & 0.95 & 3.20 & 6.36 & 0.77 & 1.00 & 2.27 & 1.88 & 2.35 & 1.95 \\
& Tyche & 1.09 & 1.12 & \textbf{1.19} & \textbf{0.93} & 0.83 & \textbf{0.69} & \textbf{2.71} & \textbf{5.17} &  \textbf{0.60} & \textbf{0.95} & 2.13 & 1.85 & 2.21 & 1.86 \\
\midrule
\multirow{3}{*}{3-Day} 
& GraphCast      & 2.21 & 2.36 & 2.28 & 2.86 & 2.73 & 2.29 & 5.46 & 10.23& 1.44 & 1.90 & \textbf{4.12} & \textbf{3.20} & 4.63 & \textbf{3.38} \\
& ArchesGen          & 2.15 & \textbf{2.22} & 1.88 & 2.71 & 2.53 & 1.94 & 5.30 & 10.30& 1.57 & 1.78 & 4.28 & 3.37 & 4.69 & 3.49 \\
& Tyche & \textbf{2.13} & 2.31 & \textbf{1.68} & \textbf{2.49} & \textbf{2.29} & \textbf{1.85} & \textbf{5.11} & \textbf{9.84} & \textbf{1.39} & \textbf{1.63} & 4.27 & 3.32 & \textbf{4.56} & 3.53 \\
\midrule
\multirow{3}{*}{7-Day} 
& GraphCast      & 3.97 & 3.76 & 3.62 & 6.64 & 6.73 & 5.34 & 7.87 & 15.28& 3.15 & 3.49 & 7.85 & \textbf{5.84} & 8.02 & 6.04 \\
& ArchesGen          & 4.25 & 4.05 & 3.35 & 6.84 & 7.65 & 6.08 & 7.81 & 14.72& 3.85 & 3.59 & 8.84 & 7.06 & 8.97 & 6.71 \\
& Tyche & \textbf{3.89} & \textbf{3.52} & \textbf{2.77} & \textbf{6.03} & \textbf{6.02} & \textbf{4.57} & \textbf{7.12} & \textbf{14.24}& \textbf{3.08} & \textbf{3.02} & \textbf{7.48} & 5.85 & \textbf{7.80} & \textbf{5.73} \\
\midrule
\multirow{3}{*}{10-Day} 
& GraphCast      & 4.94 & \textbf{4.59} & 4.50 & 10.28& 10.51& 8.37 & 9.22 & 17.88& 3.89 & 4.24 & 10.20& 7.54 & 10.09& 7.14 \\  
& ArchesGen          & 4.88 & 4.87 & 4.17 & 9.91 & 9.69 & 7.56 & 8.62 & 17.12& 4.33 & 4.33 & 10.34& 7.75 & 10.23& 7.57 \\ 
& Tyche & \textbf{4.71} & 4.64 & \textbf{3.74} & \textbf{8.58} & \textbf{8.97} & \textbf{6.94} & \textbf{8.44} & \textbf{16.53}& \textbf{3.87} & \textbf{3.94} & \textbf{9.62} & \textbf{7.31} & \textbf{9.77} & \textbf{7.00} \\
\bottomrule
\end{tabular}
}
% \vspace{-10pt}
\end{table}

\begin{table}[ht]
\centering
% \vspace{-10pt}
\caption{Comparison of latitude-weighted RMSE between ArchesGen and \ourmethod for surface and upper-air variables. Best results for each lead time are highlighted in \textbf{bold}.}
\label{tab:swift_vs_tyche_rmse}
\setlength{\tabcolsep}{4.5pt} 
\resizebox{1.0\textwidth}{!}{
\begin{tabular}{@{}ll *{4}{c} *{10}{c}@{}}
\toprule
\multirow{2}{*}{\textbf{Lead Time}} & \multirow{2}{*}{\textbf{Model}} & \multicolumn{4}{c}{\textbf{Surface Variables}} & \multicolumn{10}{c}{\textbf{Upper-Air Variables (Pressure Levels)}} \\
\cmidrule(lr){3-6} \cmidrule(l){7-16}
& & u10 & v10 & t2m & mslp & z7 & z9 & q7 & q9 & t8 & t10 & U7 & U9 & V7 & V9 \\
\midrule
\multirow{2}{*}{1-Day} 
& ArchesGen          & 0.86 & 0.92 & 1.06 & 1.02 & 0.74 & 0.65 & 2.54 & 4.99 & 0.62 & 0.83 & 1.81 & 1.48 & 1.81 & 1.46 \\
& Tyche & \textbf{0.79} & \textbf{0.83} & \textbf{0.88} & \textbf{0.71} & \textbf{0.58} & \textbf{0.50} & \textbf{2.35} & \textbf{4.69} & \textbf{0.50} & \textbf{0.73} & \textbf{1.57} & \textbf{1.36} & \textbf{1.63} & \textbf{1.35} \\
\midrule
\multirow{2}{*}{3-Day} 
& ArchesGen          & 1.88 & 1.94 & 1.70 & 2.60 & 2.36 & 1.81 & 4.40 & 8.30 & 1.42 & 1.62 & 3.76 & 2.91 & 4.10 & 3.01 \\
& Tyche & \textbf{1.63} & \textbf{1.71} & \textbf{1.32} & \textbf{2.00} & \textbf{1.83} & \textbf{1.45} & \textbf{3.90} & \textbf{7.68} & \textbf{1.14} & \textbf{1.29} & \textbf{3.29} & \textbf{2.55} & \textbf{3.59} & \textbf{2.67} \\
\midrule
\multirow{2}{*}{7-Day} 
& ArchesGen          & 3.62 & 3.15 & 2.68 & 6.02 & 6.59 & 5.09 & 6.38 & 12.02& 3.24 & 2.89 & 6.86 & 5.22 & 6.67 & 5.28 \\
& Tyche & \textbf{3.12} & \textbf{2.77} & \textbf{2.03} & \textbf{5.03} & \textbf{5.00} & \textbf{3.92} & \textbf{5.53} & \textbf{11.12}& \textbf{2.39} & \textbf{2.32} & \textbf{6.13} & \textbf{4.76} & \textbf{6.19} & \textbf{4.61} \\
\midrule
\multirow{2}{*}{10-Day} 
& ArchesGen          & 3.79 & 3.85 & 3.37 & 8.15 & 8.24 & 6.32 & 7.14 & 14.11& 3.62 & 3.58 & 7.95 & 5.93 & 8.23 & 5.98 \\
& Tyche & \textbf{3.74} & \textbf{3.68} & \textbf{3.31} & \textbf{7.29} & \textbf{7.58} & \textbf{5.80} & \textbf{6.44} & \textbf{12.53}& \textbf{3.38} & \textbf{3.38} & \textbf{7.67} & \textbf{5.83} & \textbf{7.79} & \textbf{5.73} \\
\bottomrule
\end{tabular}
}
% \vspace{-10pt}
\end{table}

\subsection{Comparison of Spread skills}
\label{app:skill_ens}
% spread可视化

\begin{figure*}[t]
\centering
\includegraphics[width=1.0\linewidth]{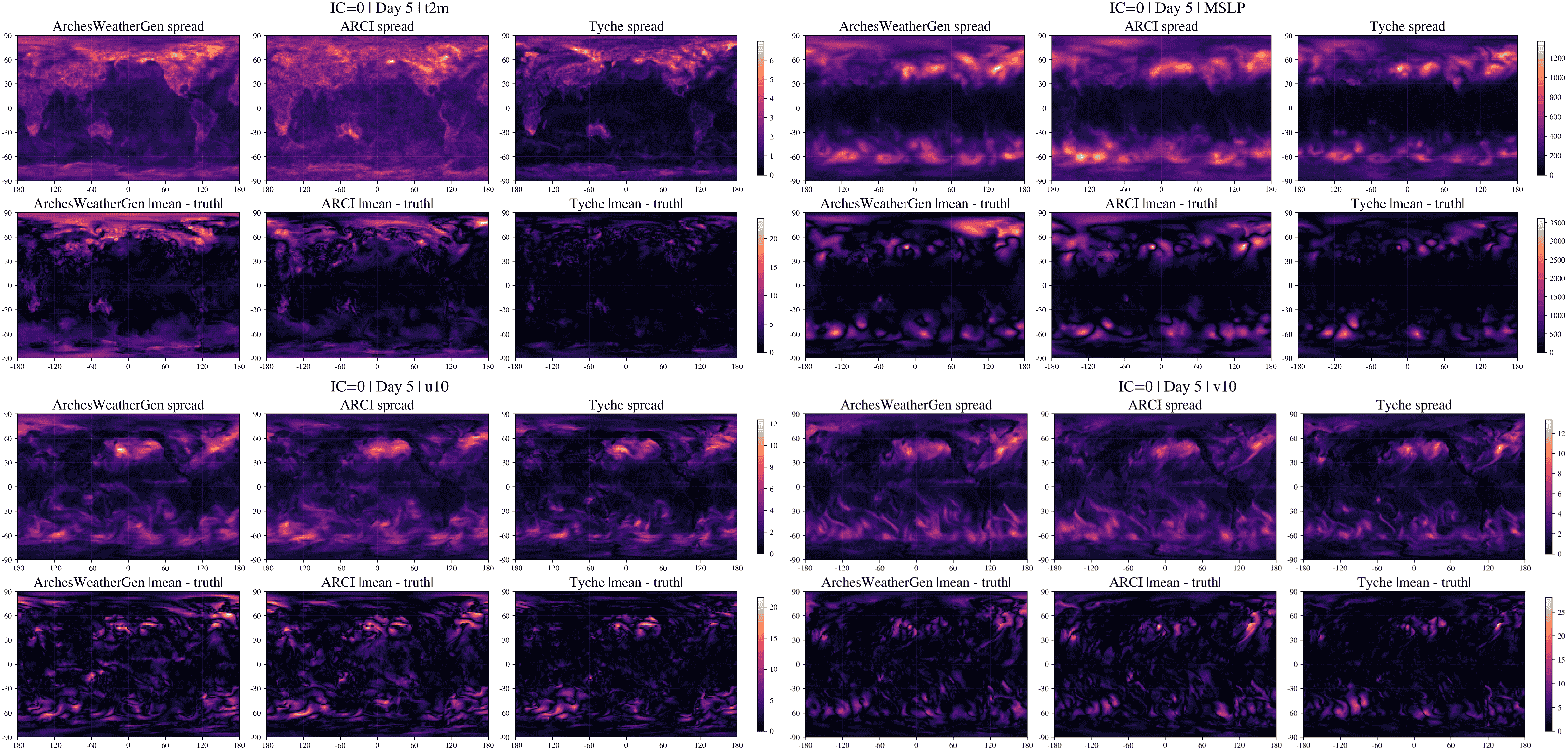}
% \vspace{-15pt}
\caption{Spatial comparison of ensemble spread (top rows) and absolute mean error (bottom rows) at a 5-day lead time.}
\label{fig:ap_spread}
% \vspace{-15pt}
\end{figure*}

To comprehensively evaluate the spatial reliability of uncertainty estimation, we conducted a comparative visual analysis between the ensemble spread and the absolute error of the ensemble mean ($|\text{mean} - \text{truth}|$). In a well-calibrated forecasting system, the spatial distribution of the spread should tightly align with the forecast error. 
Figure \ref{fig:ap_spread} visualizes this alignment at a 5-day lead time for key surface variables (t2m, mslp, u10, and v10). 

As illustrated in Figure \ref{fig:ap_spread}, \ourmethod demonstrates a precise spatial correlation between its predicted uncertainty and actual errors. It accurately captures the complex variance in dynamically active regions (\textit{e.g.}, the Southern Ocean) without suffering from pathological under-dispersion. Compared to multi-step generative baselines like ArchesWeatherGen and ARCI, \ourmethod maintains highly competitive and physically coherent uncertainty structures. This empirically validates that our CRPS-based fine-tuning effectively calibrates spatial dispersion and achieves robust reliability.

\subsection{Detailed visualizations on more horizons and variables}
\label{app:detailed_vis}

To further demonstrate the physical fidelity and temporal stability of our method, the following Figure~\ref{fig:ap_1day}, ~\ref{fig:ap_3day}, ~\ref{fig:ap_7day} present comprehensive visualizations of \ourmethod's single-member forecasts and absolute errors across nine meteorological variables at 1-day, 3-day, and 7-day lead times.

\begin{figure*}[!t]
\centering
\includegraphics[width=1.0\linewidth]{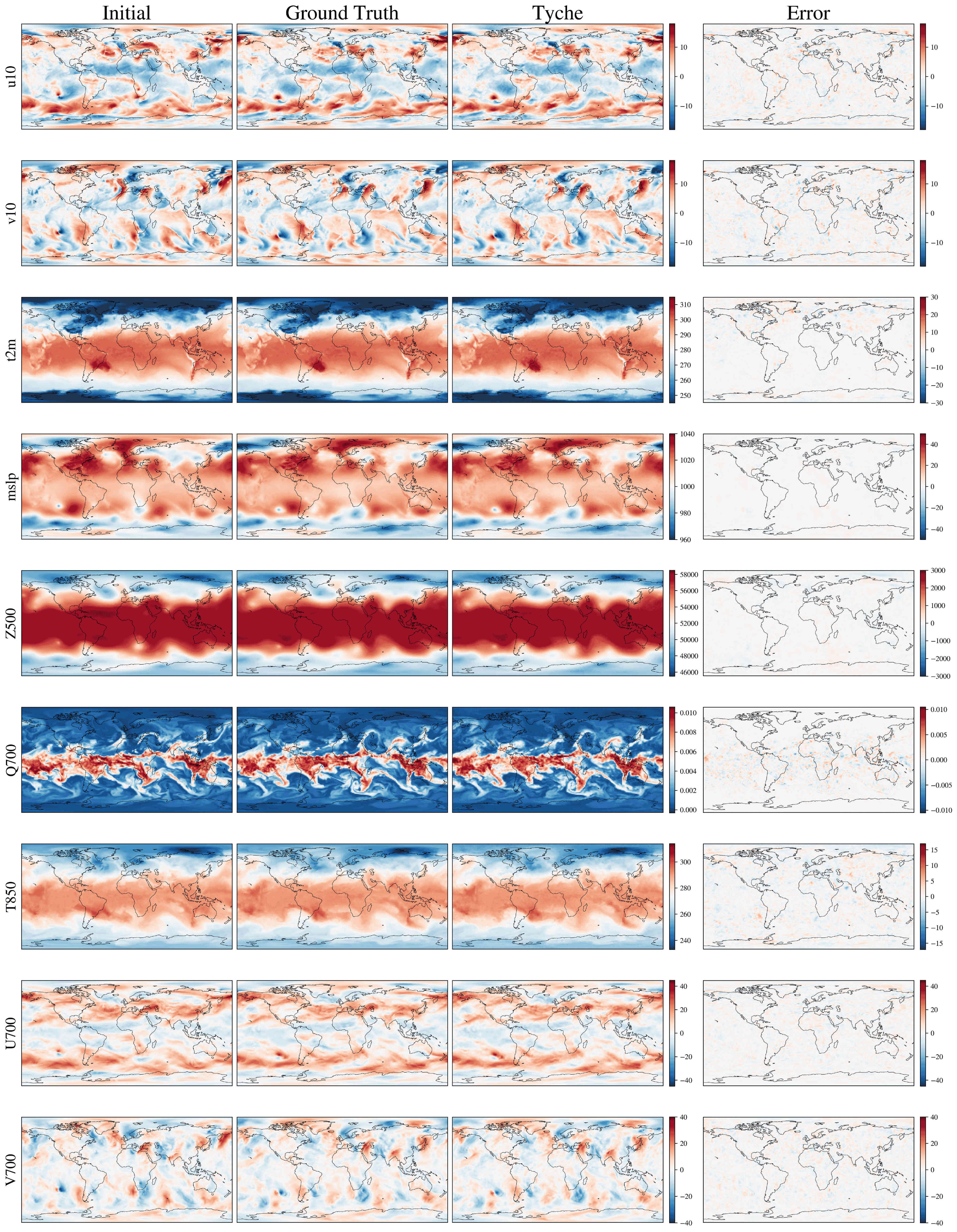}
\vspace{-15pt}
\caption{Detailed visualization of 1-day single-member forecasts for nine selected variables. 
}
\label{fig:ap_1day}
\vspace{-10pt}
\end{figure*}

\begin{figure*}[!t]
\centering
\includegraphics[width=1.0\linewidth]{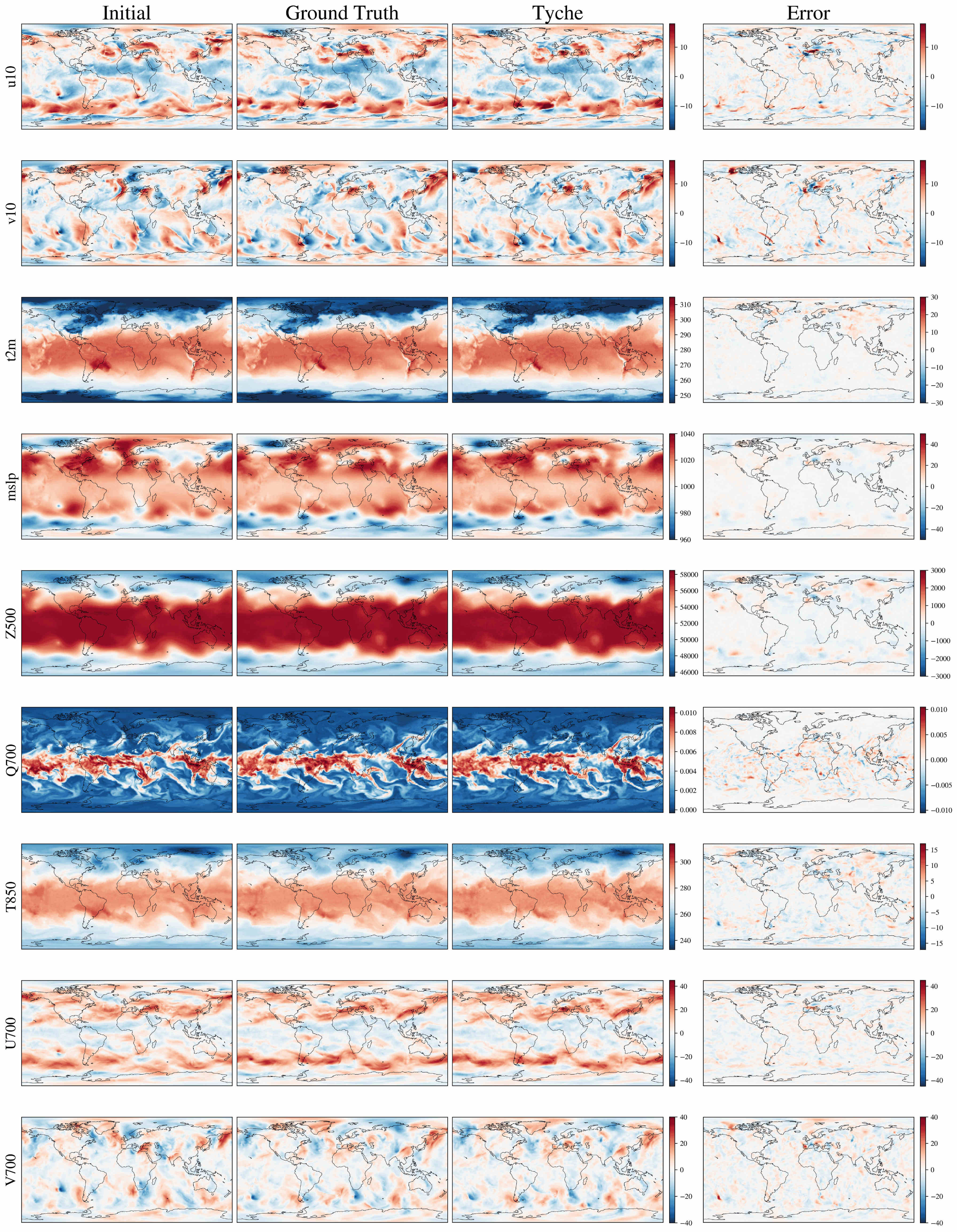}
\vspace{-15pt}
\caption{Detailed visualization of 3-day single-member forecasts for nine selected variables. 
}
\label{fig:ap_3day}
\vspace{-10pt}
\end{figure*}

\begin{figure*}[!t]
\centering
\includegraphics[width=1.0\linewidth]{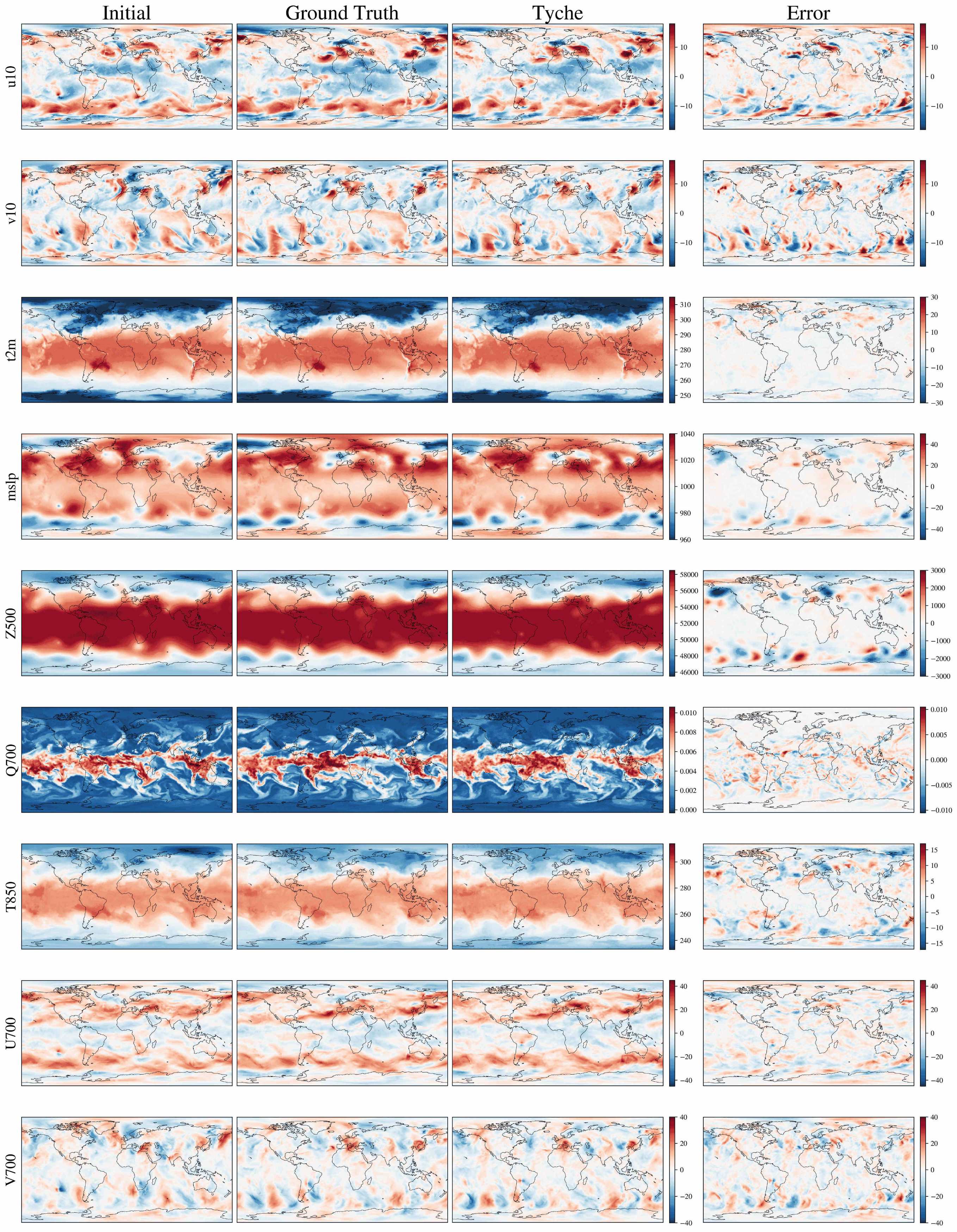}
\vspace{-15pt}
\caption{Detailed visualization of 7-day single-member forecasts for nine selected variables. 
}
\label{fig:ap_7day}
\vspace{-10pt}
\end{figure*}

\clearpage

\section*{NeurIPS Paper Checklist}

\begin{enumerate}

\item {\bf Claims}
    \item[] Question: Do the main claims made in the abstract and introduction accurately reflect the paper's contributions and scope?
    \item[] Answer: \answerYes{}
    \item[] Justification: The main claims regarding the 1-NFE efficiency, the generative transport method, and competitive performance are explicitly stated in the Abstract and Introduction (Section~\ref{introduction}), and are fully supported by theoretical analysis (Appendix~\ref{app:theory}) and empirical results (Section~\ref{experiment}).

\item {\bf Limitations}
    \item[] Question: Does the paper discuss the limitations of the work performed by the authors?
    \item[] Answer: \answerYes{}
    \item[] Justification: We have included it in the conclusion.

\item {\bf Theory assumptions and proofs}
    \item[] Question: For each theoretical result, does the paper provide the full set of assumptions and a complete (and correct) proof?
    \item[] Answer: \answerYes{}
    \item[] Justification: Theorem and rigorous mathematical proofs are formally provided in Appendix~\ref{app:theory}.

    \item {\bf Experimental result reproducibility}
    \item[] Question: Does the paper fully disclose all the information needed to reproduce the main experimental results of the paper to the extent that it affects the main claims and/or conclusions of the paper (regardless of whether the code and data are provided or not)?
    \item[] Answer: \answerYes{}
    \item[] Justification: We provide comprehensive details regarding the dataset preprocessing (Appendix \ref{app:data}), baseline configurations (Appendix \ref{app:baseline}, network architecture (Appendix \ref{app:tyche}) and training pipeline (Appendix \ref{app:implementation_details}), and evaluation metrics (Appendix ~\ref{app:metrics}) to fully facilitate reproducibility.

\item {\bf Open access to data and code}
    \item[] Question: Does the paper provide open access to the data and code, with sufficient instructions to faithfully reproduce the main experimental results, as described in supplemental material?
    \item[] Answer: \answerYes{}
    \item[] Justification: The project repository is linked in the abstract, and the ERA5 dataset and WeatherBench2 are publicly available. We have provided detailed implementation in Appendix~\ref{app:baseline}.

\item {\bf Experimental setting/details}
    \item[] Question: Does the paper specify all the training and test details (e.g., data splits, hyperparameters, how they were chosen, type of optimizer) necessary to understand the results?
    \item[] Answer: \answerYes{}
    \item[] Justification: Detailed experimental settings, including data splits (years 1979-2017 for training, 2020 for testing), optimization hyperparameters, curriculum stages, and architecture specifics, are extensively documented in Section~\ref{exp:setting} and Appendix~\ref{app:setting}.

\item {\bf Experiment statistical significance}
    \item[] Question: Does the paper report error bars suitably and correctly defined or other appropriate information about the statistical significance of the experiments?
    \item[] Answer: \answerYes{}
    \item[] Justification: Following standard practices in global ML weather forecasting (e.g., WeatherBench2), we evaluate performance by aggregating metrics (RMSE, CRPS) over a massive number of spatial grid points and a full year of temporal test data (2020), which naturally stabilizes the results. 

\item {\bf Experiments compute resources}
    \item[] Question: For each experiment, does the paper provide sufficient information on the computer resources (type of compute workers, memory, time of execution) needed to reproduce the experiments?
    \item[] Answer: \answerYes{}
    \item[] Justification: We explicitly state the hardware resources required for training (eight 80GB NVIDIA H100 GPUs) and inference (a single GPU) in Section~\ref{exp:setting} and Appendix~\ref{app:implementation_details}.
    
\item {\bf Code of ethics}
    \item[] Question: Does the research conducted in the paper conform, in every respect, with the NeurIPS Code of Ethics \url{https://neurips.cc/public/EthicsGuidelines}?
    \item[] Answer: \answerYes{}
    \item[] Justification: This research focuses on improving atmospheric forecasting using publicly available scientific data, raising no ethical concerns, and strictly conforms to the NeurIPS Code of Ethics.

\item {\bf Broader impacts}
    \item[] Question: Does the paper discuss both potential positive societal impacts and negative societal impacts of the work performed?
    \item[] Answer: \answerYes{}
    \item[] Justification: The potential positive societal impacts of the work, such as improving high-impact decision-making in aviation, energy, agriculture, and disaster response, are discussed in the Introduction.
    
\item {\bf Safeguards}
    \item[] Question: Does the paper describe safeguards that have been put in place for responsible release of data or models that have a high risk for misuse (e.g., pre-trained language models, image generators, or scraped datasets)?
    \item[] Answer: \answerNA{}
    \item[] Justification: The paper focuses on weather forecasting models based on physical atmospheric data, which do not pose high risks for malicious misuse (e.g., generation of deceptive content).

\item {\bf Licenses for existing assets}
    \item[] Question: Are the creators or original owners of assets (e.g., code, data, models), used in the paper, properly credited and are the license and terms of use explicitly mentioned and properly respected?
    \item[] Answer: \answerYes{}
    \item[] Justification: We properly cite the original papers for the ERA5 dataset and the WeatherBench2 evaluation framework, respecting their respective open-access terms.

\item {\bf New assets}
    \item[] Question: Are new assets introduced in the paper well documented and is the documentation provided alongside the assets?
    \item[] Answer: \answerNA{}
    \item[] Justification: The paper introduces a novel methodology and architecture but does not release new benchmark datasets.

\item {\bf Crowdsourcing and research with human subjects}
    \item[] Question: For crowdsourcing experiments and research with human subjects, does the paper include the full text of instructions given to participants and screenshots, if applicable, as well as details about compensation (if any)? 
    \item[] Answer: \answerNA{}
    \item[] Justification: The research does not involve crowdsourcing or human subjects.

\item {\bf Institutional review board (IRB) approvals or equivalent for research with human subjects}
    \item[] Question: Does the paper describe potential risks incurred by study participants, whether such risks were disclosed to the subjects, and whether Institutional Review Board (IRB) approvals (or an equivalent approval/review based on the requirements of your country or institution) were obtained?
    \item[] Answer: \answerNA{}
    \item[] Justification: The research does not involve human subjects.

\item {\bf Declaration of LLM usage}
    \item[] Question: Does the paper describe the usage of LLMs if it is an important, original, or non-standard component of the core methods in this research? Note that if the LLM is used only for writing, editing, or formatting purposes and does \emph{not} impact the core methodology, scientific rigor, or originality of the research, declaration is not required.
    %this research? 
    \item[] Answer: \answerNA{}
    \item[] Justification: The core method development in this research does not involve LLMs as any component. 

\end{enumerate}

\end{document}